\pdfoutput=1
\PassOptionsToPackage{backref=page}{hyperref}
\PassOptionsToPackage{numbers, compress}{natbib}
\documentclass{article}

\usepackage[T1]{fontenc}
\usepackage{microtype}
\usepackage{graphicx}
\usepackage{subfigure}
\usepackage{booktabs}
\usepackage{tabularx}
\usepackage{enumitem}
\usepackage{algorithm}
\usepackage{algorithmic}
\usepackage{wrapfig}

\usepackage{hyperref}
\newcommand{\theHalgorithm}{\arabic{algorithm}}

\usepackage[final]{neurips_glfrontiers_2023}

\usepackage{balance}
\usepackage{amsmath}
\usepackage{amssymb}
\usepackage{mathtools}
\usepackage{amsthm}
\usepackage{mathtools}  \usepackage{nicefrac}
\usepackage{xspace}

\usepackage{tikz}
\usetikzlibrary{fit,positioning,calc,shapes,backgrounds}
\usepackage{pgfplots}
\pgfplotsset{compat=1.16}
\usepgfplotslibrary{fillbetween, groupplots}

\usepackage{todonotes} \newcommand{\bryan}[1]{\todo[inline,color=green!20!white]{\textbf{bryan:} #1}}
\newcommand{\anton}[1]{\todo[inline,color=blue!20!white]{\textbf{anton:} #1}}
 \theoremstyle{plain}
\newtheorem{theorem}{Theorem}[section]

\theoremstyle{definition}
\newtheorem{definition}[theorem]{Definition}

\theoremstyle{remark}

\DeclareMathOperator{\tr}{Tr}

\DeclareRobustCommand{\bigO}{\ifmmode
         \mathcal{O}
    \else
        \GenericError{}{Attempt to use bigO outside of math mode}\fi
}
\makeatother

\DeclareRobustCommand{\bigpolyO}{\ifmmode
         \Tilde{\mathcal{O}}
    \else
        \GenericError{}{Attempt to use bigO outside of math mode}\fi
}
\makeatother

\renewcommand*{\backrefalt}[4]{\ifcase #1 No citations.\or
Cited on page #2.\else
Cited on pages #2.\fi
}

\newcommand*{\mA}{\mathbf{A}}
\newcommand*{\mD}{\mathbf{D}}
\newcommand*{\mI}{\mathbf{I}}
\newcommand*{\mL}{\mathbf{L}}
\newcommand*{\mLn}{\tilde{\mathbf{L}}}
\newcommand*{\mP}{\mathbf{P}}

\newcommand*{\sR}{\mathbb{R}}
\newcommand*{\sN}{\mathbb{N}}

\definecolor{cycle1}{RGB}{235,172,35}
\definecolor{cycle2}{RGB}{184,0,88}
\definecolor{cycle3}{RGB}{0,140,249}
\definecolor{cycle4}{RGB}{0,110,0}
\definecolor{cycle5}{RGB}{0,187,173}
\definecolor{cycle6}{RGB}{209,99,230}
\definecolor{cycle7}{RGB}{178,69,2}
\definecolor{cycle8}{RGB}{255,146,135}
\definecolor{cycle9}{RGB}{89,84,214}
\definecolor{cycle10}{RGB}{0,198,248}
\definecolor{cycle11}{RGB}{135,133,0}
\definecolor{cycle12}{RGB}{0,167,108}
\definecolor{cyclegray}{RGB}{189,189,189}

\newtheoremstyle{break}
  {\topsep}{\topsep}{\itshape}{}{\bfseries}{}{\newline}{}\theoremstyle{break}
\newtheorem{breakhypo}{Hypothesis}

\newcommand*{\thiswork}{kTree\xspace}

\newcommand*{\glottery}{GLT\xspace}

\begin{document}
\iftrue
\author{Anton Tsitsulin \\ Google Research\And Bryan Perozzi \\ Google Research}
\title{The Graph Lottery Ticket Hypothesis: \\ Finding Sparse, Informative Graph Structure}
\maketitle

 \begin{abstract} Graph learning methods help utilize implicit relationships among data items, thereby reducing training label requirements and improving task performance.
However, determining the optimal graph structure for a particular learning task remains a challenging research problem.

In this work, we introduce the Graph Lottery Ticket (\glottery) Hypothesis -- that there is an extremely sparse backbone for every graph, and that graph learning algorithms attain comparable performance when trained on that subgraph as on the full graph.
We identify and systematically study 8 key metrics of interest that directly influence the performance of graph learning algorithms. 
Subsequently, we define the notion of a ``winning ticket'' for graph structure -- an extremely sparse subset of edges that can deliver a robust approximation of the entire graph's performance.
We propose a straightforward and efficient algorithm for finding these GLTs in arbitrary graphs.
Empirically, we observe that performance of different graph learning algorithms can be matched or even exceeded on graphs with the average degree as low as 5.
\end{abstract} \section{Introduction}\label{sec:introduction}

Graph data naturally arises in many domains, including social
networks, interactions on the Web, and in many biological applications.
Building graphs directly from data proves useful for massive-scale data analysis; for instance, graphs can be clustered in near-linear time~\cite{dhulipala2021hierarchical}.

\begin{wrapfigure}[16]{R}{0.4\textwidth}
    \centering
    \vspace{-24pt}
    \includegraphics[width=\linewidth]{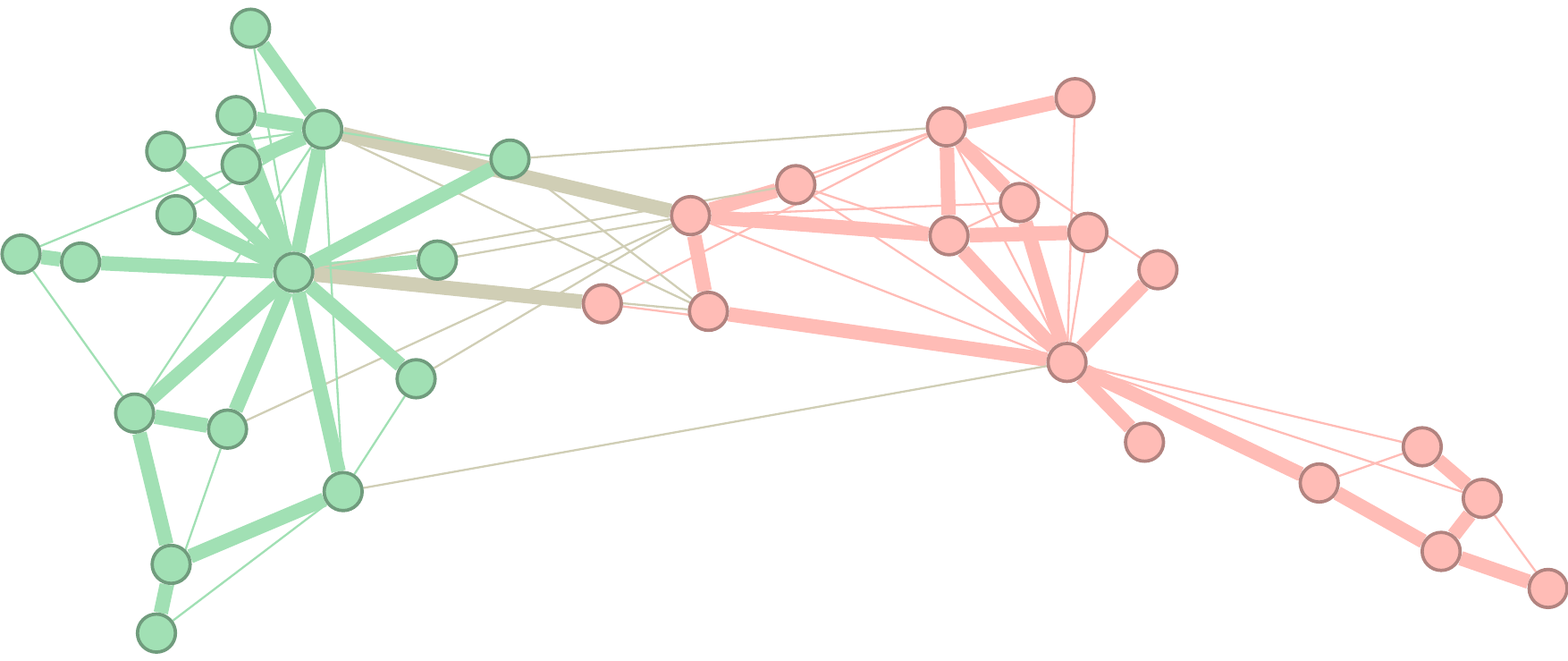}
    \vspace{-24pt}
    \caption{The Graph Lottery Hypothesis postulates that there is a sparse substructure (a \emph{winning ticket}) present in all graphs which captures its utility for graph learning tasks.
    The winning ticket of the Karate club graph~\cite{zachary1977information} in bold.}
    \label{fig:my_label}
\end{wrapfigure}

In recent years, graph machine learning has become a dominant paradigm in analysis of network data.
The performance of many graph learning algorithms is heavily dependent on the structure of data in terms of the graph curvature~\cite{topping2022understanding,seshadhri2020impossibility}, intrinsic dimensionality~\cite{tsitsulin2019spectral}, or many other metrics \cite{palowitch2022graphworld}.
A natural compulsion is to rewire graphs to optimize such metrics.
However, adding or rewiring edges may hallucinate connections that could never exist -- violating the natural graph structure.

In this paper, we investigate the general problem of finding sparse subgraphs well-suited for graph learning -- \emph{graph lottery tickets}.  
Unlike most existing work, we focus on finding substructures \emph{already present} in data, just like the ``lottery tickets'' in deep neural networks parameters~\cite{frankle2019lottery}.
We briefly formalize this notion as follows:
\begin{breakhypo}[\textbf{Graph Lottery Ticket Hypothesis}]
Any graph contains a sparse subset of edges that---when trained on that subset only---any graph learning algorithm can match the performance of the original graph.
\end{breakhypo}

We summarize our key contributions as follows:
\begin{itemize}[noitemsep,topsep=-8pt,parsep=0pt,partopsep=0pt]
    \item We formulate the Graph Lottery Ticket (\glottery) hypothesis that implies the existence of an extremely sparse backbone for every graph for which graph learning algorithms attain comparable performance as on the full graph.
    \item We propose a straightforward yet efficient algorithm to recover ``winning tickets'' -- extremely sparse subgraphs which still preserve task performance.
    \item Our experimental results illustrate our method's effectiveness. The winning tickets (sparse networks) we find match the performance for three graph learning algorithms, but with much lower average degree ($\approx$ 5).
\end{itemize}

 \section{Preliminaries and Related Work}\label{sec:related-work}

This section reviews previous attempts to optimize the structure of graphs for graph learning tasks including approaches that change the graph structure implicitly.
Before diving into the related work, Section~\ref{ssec:rw:preliminaries} establishes basic notation to be used throughout the paper.

\subsection{Preliminaries}\label{ssec:rw:preliminaries}

A graph is a pair $G = (V, E)$ with $n$ vertices $V = (v_1,\cdots ,v_n)$, $\lvert{}V\rvert = n$, and edges $E \subseteq V \times V, \lvert{}E\rvert = m$, represented by an adjacency matrix $\mA$ for which $\mA_{ij} = 1$ if $e_{ij} \in E$\footnote{For readability purposes we use ``node $i$'' instead of $v_i$ here and further, wherever appropriate.} is an edge between nodes $i$ and $j$, otherwise $\mA_{ij} = 0$.
We denote the neighborhood set of the node $u$ as $N(u)={v : (u, v)\in E}$.
Then, $\#_\Delta(i,j)=N(i)\cap{}N(j)$ denotes the set of triangles with the edge $(i,j)$.
For generality and simplicity of notation, we assume undirected and unweighted graphs, however, content of the paper can be easily generalized to the weighted and directed cases.

The degree of a node is defined as $d_i = \lvert{}N(i)\rvert$, and the
degree matrix $\mD$ is the diagonal matrix with node degrees $\mD_{ii} = d_i$.
The combinatorial (unnormalized) Laplacian matrix of a graph is defined as $\mL=\mD-\mA$.
Its normalized counterpart $\mLn$ is defined as $\mLn=\mI-\mD^{\nicefrac{-1}{2}}\mA\mD^{\nicefrac{-1}{2}}$, where $\mI$ is the identity matrix.
We use $(\lambda_1, \cdots, \lambda_n)$ to denote the ordered set of eigenvalues of graph Laplacians and $(\mu_1, \cdots, \mu_n)$ -- of graph adjacency.

\subsection{Graph Sparsifiers and Spanners}\label{ssec:rw:sparsification}

Graph sparsifier is a sparse subgraph that preserves particular properties of the original graph.
For instance, the surprising fact that $\varepsilon$-approximate cut sparsifier with $\bigpolyO(\nicefrac{n}{\varepsilon^2})$ edges can be constructed in $\bigpolyO(m)$ time was first established in 
\cite{karger1994using,benczur1996approximating}.
That notion was strengthened \cite{spielman2011spectral} to \emph{spectral sparsifiers} -- a graph $\Tilde{G}$ is called a spectral sparsifier of $G$ if
\begin{equation*}\label{eq:spectral-sparsifier}
(1-\varepsilon) x^\top\mL_{\Tilde{G}}x \leq x^\top\mL_Gx \leq (1+\varepsilon) x^\top\mL_{\Tilde{G}}x
\end{equation*}
for all $x\in\sR^V$.
Cut sparsifiers are only required to satisfy these inequalities for all
$x \in \{0, 1\}^V$.
The factors hidden in $\bigpolyO$ are, however, large.
Good sparsifiers, e.g.~\cite{spielman2008graph,batson2009twice}, are computationally expensive, limiting their practicality.
More scalable solutions, e.g.~\cite{fung2011general}, are restricted to cut sparsification and do not guarantee graph connectivity, which is crucial for many graph learning algorithms.

Spanners~\cite{peleg1989graph} provide a combinatorial view to sparsification.
Instead of preserving algebraic properties of linear systems, spanners preserve the distances in graphs with multiplicative ($t$-spanners) or additive ($+\beta$-spanners) distortion.
\cite{althofer1993sparse} propose to find $t$-spanners via a generalization of the classical greedy minimum spanning tree algorithm due to~\cite{kruskal1956shortest}.
\cite{carey2022stars} proposes a way to sparsify near-cliques during graph construction process.
The modified graph is provably a 2-hop spanner of the original, however, the number of spurious added edges can be of size of the graph itself.
In general, it is unclear how graph distances translate to the performance of graph learning algorithms.

\subsection{Graph Rewiring}\label{ssec:rw:rewiring}

Graph rewiring approaches aim to optimize the structure of a given graph via changing, adding, or deleting edges.
A heuristic edge-swap algorithm was proposed in~\cite{chan2016optimizing} to optimize multiple spectral graph robustness measures (which we review in Section~\ref{sec:metrics}) with updates computed using matrix perturbation theory.
The same strategy is used in~\cite{karhadkar2022fosr} with an even more crude update approximation for improving the algebraic connectivity, leading to improvements in learning graph neural networks.
In a similar vein, \cite{topping2022understanding} propose a greedy rewiring algorithm for optimizing the structure of a graph for a modified definition of augmented Forman curvature.
A different optimization metric was offered by~\cite{banerjee2022oversquashing}: they flip edges that minimize the number of triangles in a graph.
These methods introduces spurious edges to the graph and keep the total number of edges approximately the same.
Similarly,~\cite{chen2021unified} proposes to sparsify a graph iteratively with training a GNN model.
In contrast, this works finds extremely sparse subgraphs \emph{without} spurious edges and in a model-agnostic fashion.

Contrapositively, \cite{klicpera2019diffusion} propose to augment the edges of the graph with extra edges derived from the diffusion process from the original graph.
This approach densifies the graph to an extreme degree, sometimes adding hundred times more edges than in the original graph.

\subsection{Implicit Graph Rewiring}\label{ssec:rw:implicit}

Many graph learning methods implicitly modify the graph to achieve scalability linear in terms of the number of nodes.
A common approach for scaling up GNN training to large graphs is to sample rooted subgraphs from each node~\cite{hamilton2017inductive,chen2018fastgcn}.
While graph that were implicitly sampled during GNN training have constant degree in theory, the upper bound, assuming parameters from~\cite{hamilton2017inductive}, is 2500 neighbors per node, which significantly densifies the graph.
In another vein,~\cite{arnaiz2022diffwire} propose to rewire the subgraphs \emph{during} GNN training to optimize the connectivity of these sampled subgraphs.
This approach densifies local subgraphs and is not applicable to general graph learning algorithms.

The same is true for sampling in the process of graph embedding.
DeepWalk~\cite{perozzi2014deepwalk} samples long random walks from each node, and further densifies the implicit graph by running a long-range window
An example more amenable for analysis is the sampling process of personalized PageRank-based embedding methods, e.g.~\cite{tsitsulin2018verse}.
Even with approximate computation~\cite{andersen2007using}, PPR values of the neighborhood nodes are $\bigO(\alpha(1-\alpha))\gg0$, meaning the graph is densified to an extreme degree.
 \section{What is a Good Graph Structure?}\label{sec:metrics}

Structural graph properties have an outsized impact on the performance of graph learning algorithms, however, to our knowledge, there is no systematic study of the phenomenon.
This section covers that from two different perspectives on graph structure: spectral expansion properties and local edge curvature.
Through these two lenses we try to answer the question in the section title---what does make graph structure good?

\subsection{Spectral Properties}

Laplacian systems are at the heart of many graph machine learning, including label propagation~\cite{zhu2003semi}, clustering~\cite{ng2001spectral}, and more.
Condition number $\kappa(\mA)=\frac{\lambda_n}{\lambda_1}$ bounds the convergence rate of iterative algorithms for solving linear equations in $\mA$.
Since graph Laplacians are singular, the convergence can be instead measured in terms of the finite condition number $\kappa_f=\nicefrac{\lambda_n}{\lambda_2}$.
From a signal propagation perspective, $\lambda_2$ is related to the worst-case mixing of a random walk over $G$.

Algebraic connectivity, the second eigenvalue of the graph Laplacian, is ubiquitous due to its relation to vertex connectivity.
For instance, $\lambda_2\geq\frac{4}{nD}$, where $D$ is graph's diameter, but the most exciting appearance of $\lambda_2$ is arguably in the Cheeger constant $h(G)$ of a graph,
which is the lowest-density cut of the graph normalized by cut size.
Algebraic connectivity can be used to bound the Cheeger constant~\cite{chung1997spectral}: $\frac{\lambda_2}{2} \leq h(G) \leq \sqrt{2\lambda_2}$.

Over-smoothing in GNNs happens with the rate of  $\bigO((s\lambda_2)^L)$, where $s$ is the largest singular value of node features and $L$ is the number of GNN layers~\cite{oono2020graph,cai2020note}.
While high oversmoothing does not sound very desirable, \cite{karhadkar2022fosr} showed that relational GCNs are \emph{flexible} in how much the smooth the graph, in the range of $[0, \lambda_2]$, as measured by the Dirichlet energy of the GCN layer.
Therefore, having large algebraic connectivity should be considered advantageous from graph neural network perspective.

High $\lambda_2$ implies that a graph can not be well embedded in $\sR$~\cite{ghosh2006growing}. 
For higher-dimensional Euclidean embeddings, \cite{tsitsulin2019spectral} empirically studies the reconstruction ability with respect to the spectral dimensionality of graphs.
Instead of computing the spectral dimensionality directly, they estimate the graph Laplacian eigenvalue growth rate.
While it may be easier to embed graphs with small $\lambda_2$, we are interested in the most informative subgraphs of a given graph.
Therefore, evidence from both GNNs and graph embedding points to positive effects for maximizing $\lambda_2$, which we study in Section~\ref{ssec:experiments-stats}.

Graph robustness studies~\cite{costa2007characterization} introduced two additional spectral measures.
Spectral radius---the largest eigenvalue of the adjacency matrix---controls the speed of various dynamic processes defined on graphs, for instance, the spread of contagious viruses.
Total number of spanning trees can be thought of as the total number of ways information can be transmitted in the network.
Due to the matrix-tree theorem, it can be efficiently approximated as a product of the eigenvalues of the graph Laplacian.
We use both spectral radius and the number of spanning trees in our experimental study.

\subsection{Curvature}\label{ssec:metrics:curvature}

Graph curvature~\cite{forman2003bochner,ollivier2009ricci} adapts the notion of ``flatness'' from manifolds to graphs.
Near-cliques tend to have large positive curvature, planar grids have zero curvature, and trees have negative curvature.
Forman curvature is the most computationally efficient version that is also easier to analyze combinatorially.
There are multiple definitions of Forman curvature, we introduce the one due to~\cite{samal2018comparative}, since it was shown that augmented Forman curvature is tightly correlated with definition due to \cite{ollivier2009ricci}.
\begin{definition}
For any edge $(i,j)$ the augmented Forman Ricci curvature is given by
$$
F^\#(i,j) = 4 - d_i - d_j + 3 \gamma \lvert\#_\Delta(i,j)\rvert, \qquad \gamma > 0.
$$
\end{definition}

An exciting recent development~\cite{devriendt2022discrete} connects the notion of the \emph{effective resistance} to curvature of graphs.
Effective resistance is defined through the Moore-Penrose pseudoinverse of the graph Laplacian $\mL^\dagger$ as $\omega(i,j) = (e_i-e_j)^\top{}\mL^\dagger(e_i-e_j)$.
\begin{definition}
For a node $i$, the link resistance curvature is given by
$
\rho_i = 1 - \frac{1}{2} \sum_{j\in{}N(i)}{\omega(i, j)}.
$
\end{definition}

All notions of curvature have intimate connections to the number of triangles.
Effective resistance of an edge is bounded by the number of triangles containing this edge:
$\omega(i,j)\leq\frac{2}{\#_\Delta(i,j)+2}$.
\cite{seshadhri2020impossibility} proves that it is impossible to faithfully embed triangle-rich graphs in the Euclidean space\footnote{\cite{chanpuriya2020node} shows how nonlinear embedding models are able to circumvent this restriction.}.
This provides evidence against having too many triangles in the graph for faithful embedding.

There is evidence~\cite{topping2022understanding} that large negative curvature leads to over-squashing of the gradients in graph neural networks.
However, negative negative curvature is not strictly bad for GNNs -- \cite{deac2022expander} shows how propagating the information alongside the edges of a random expander graph with small negative curvature empirically improves performance of GNNs.

These results in graph curvature motivate us to include a scalable approximation~\cite{ubaru2017fast} to the total number of triangles in a graph and its total effective resistance $R=\sum_{i,j\in{}E}\omega(i,j)=n\sum_i\lambda_i^{-1}$ as metrics in experiments in Section~\ref{ssec:experiments-stats}.
Additionally, we include a bound~\cite{jost2014ollivier} on the Ollivier's notion of curvature by the means of local graph clustering coefficient of~\cite{watts1998collective}.
In total, we will experimentally study three metrics related to graph curvature. \section{Finding Winning Graph Lottery Tickets}\label{sec:method}

\begin{algorithm}[t]
	\begin{algorithmic}[1]\caption{\textsc{kTree}$(G,\Bar{m})$}\label{alg:main}
	\STATE \textbf{Input:} Graph $G$, target number of edges $\Bar{m}$.
	\STATE \textbf{Output:} \glottery of $G$.
	\STATE $GLT \gets (V, \emptyset)$
	\WHILE {$\lvert E_{GLT}\rvert\leq{}\Bar{m}$}
        \STATE $T\gets \textsc{RandomTree}(G)$.
        \IF {$\lvert E_{GLT}\rvert\leq\Bar{m}-n+1$}
            \STATE $GLT \gets GLT \cap T$
        \ELSE
            \STATE $GLT \gets \textsc{RandomSelect}(T, \Bar{m}-\lvert E_{GLT}\rvert)$
        \ENDIF
    \ENDWHILE
    \STATE Output $GLT$.
	\end{algorithmic}
\end{algorithm}
\begin{algorithm}[t]
	\begin{algorithmic}[1]\caption{\textsc{1Tree}$(G,\Bar{m})$}\label{alg:1tree}
	\STATE \textbf{Input:} Graph $G$, target number of edges $\Bar{m}$.
	\STATE \textbf{Output:} \glottery of $G$.
	\STATE $GLT \gets \textsc{RandomTree}(G)$
	\STATE $GLT \gets\textsc{RandomSelect}(E_G, \Bar{m}-n+1)$
    \STATE Output $GLT$.
	\end{algorithmic}
\end{algorithm}

As we can see from the previous section, there is no single metric dictating performance of graph learning algorithms.
Therefore, a one-size-fits-all algorithm that can produce graph lottery tickets that optimize all the metrics simultaneously does not exist.
Instead, this section presents two straightforward yet effective approaches to finding lottery ticket structure in general graphs in a scalable and effective way, which approximately optimize the metrics discussed above.

We want to stress that our formulation of \glottery does not require knowledge of \emph{which} graph learning algorithm will be run on the graph nor any extra information such as node features or labels.
Additionally, being algorithm-agnostic implies that a successful \glottery search algorithm must preserve graph connectivity, since most graph learning algorithms rely on that notion.

These requirements naturally leads us to the notion of \textit{spanning trees}.
Specifically, we propose to take a union of $k$ random spanning trees as our \glottery construction.
This approach was used to construct expander graphs and spectral sparsifiers in~\cite{goyal2009expanders}.
Algorithms~\ref{alg:main} presents the version that we use in our experiments.
Given an edge budget $\Bar{m}$, we iteratively combine random spanning trees of $G$ to form the \glottery graph.
We also experimentally study a more bare-bone version, 1Tree, which constructs a \emph{single} random spanning tree and adds random edges of $G$ to that tree (cf.~Algorithm~\ref{alg:1tree}).

There are many exciting connections of random spanning trees to various properties of graphs, mainly through the algebraic lens of the matrix-tree theorem.
One of the most interesting connections is to the notion of the effective resistance: the probability of the edge being included in a random spanning tree is in fact equal to its effective resistance.

\begin{theorem}[\cite{goyal2009expanders}]
The union of two  random spanning trees of the complete graph on $n$ vertices has constant vertex expansion with probability $1-o(1)$.
\end{theorem}

Random trees were recently used as graph sparsifiers~\cite{fung2010graph}.
They show that a slightly advanced version (with extra edge reweighting step) of the Algorithm~\ref{alg:main} produces a spectral sparsifier in the sense of Equation~\ref{eq:spectral-sparsifier}.
Constructing a random spanning tree takes near-linear $\bigO(m^{1+o(1)})$ time in terms of the number of edges $m$, due to a recent algorithm due to~\cite{schild2018almost}.
Therefore, both kTree and 1Tree are almost linear in the number of the edges of the input graph.
In the next section we show that in addition to attractive computational properties, both kTree and 1Tree provide significant improvements on graph learning metrics studied in Section~\ref{sec:metrics}. \section{Experiments}\label{sec:experiments}

We present a wide range of experiments on real and synthetic graphs using (arguably) the three most popular graph learning algorithms:
\begin{itemize}[noitemsep,topsep=0pt,parsep=0pt,partopsep=0pt]
    \item Louvain graph clustering~\cite{blondel2008fast} greedily partitions the input graph hierarchically optimizing the modularity of the graph.
    \item DeepWalk graph embedding~\cite{perozzi2014deepwalk} trains a shallow neural network on a dataset of short random walks to extract node embeddings in $\sR^d$.
    \item Graph convolutional networks~\cite{kipf2017semi} uses the graph structure to propagate information for making graph-informed predictions.
\end{itemize}

In each experiment, we sparsify a a graph and run analyses on the sparse graph backbone.
Since some of our metrics depend on the total number of edges in the graph, we use a fixed number of edges corresponding to a target average node degree from the range $[1.1, 10]$.
Some graphs in our studies have an average node degree of less that 10 naturally, in this case, we stop at that number.

\subsection{Baselines}\label{sec:baselines}

We evaluate against two state-of-the-art baselines:
\begin{itemize}[noitemsep,topsep=0pt,parsep=0pt,partopsep=0pt]
    \item \textbf{Spectral radius}~\cite{chan2016optimizing,karhadkar2022fosr}: each edge is weighted as the gradient the spectral radius of the adjacency matrix of a graph.
    \item \textbf{Edge significance}~\cite{dianati2016unwinding} computes statistical edge significance for every edge. We note that for undirected and unweighted graphs this weighting strategy is equivalent to computing the contribution of an edge to the modularity metric~\cite{newman2006modularity}.
\end{itemize}

Most graph learning algorithms require input graph to be connected, moreover, some of the metrics introduced in Section~\ref{sec:metrics} are sensitive to the number of connected components in graphs.
Because of that, we slightly modify competing methods to first find a minimum spanning tree of a graph with respect to the weights produced by respective baseline, and then greedily add remaining edges.
For graph learning algorithms that are not sensitive to disconnected components we additionally report results of a completely \textbf{random} baseline.
We do not report graph-level statistics for that strategy, as many of the metrics are not defined for disconnected graphs.

\subsection{Datasets}\label{sec:datasets}

We evaluate the proposed search method on a wide selection of 7 natural graphs, 3 graphs constructed from the data, and a set of synthetic stochastic blockmodel (SBM) graphs~\cite{nowicki2001estimation}.
We provide a brief description of real-world datasets in the Appendix~\ref{sec:appendix-datasets}.
We randomize the train and test splits using the strategy of~\cite{shchur2018pitfalls} and pick 20 nodes per class as a training set, and leave all other nodes for testing. 

SBM is a generative graph model which divides graph vertices into $k$ classes, and then places edges between two vertices $i$ and $j$ with probability $p_{ij}$ derived from the assignments.
Specifically, each vertex $i$ is given a class $y_i\in\{1,\ldots, k\}$, and an edge $(i, j)$ is added with probability $\mP_{y_iy_j}$, where $\mP$ is a symmetric $k\times k$ matrix containing the between/within-community edge probabilities.
We set $\mP_{y_iy_j}=q$ if $i=j$ and to $p$ otherwise.
In this simple setup, $\nicefrac{p}{q}$ is the signal-to-noise ratio that measures the strength of the assortativity of a graph.
For our graph statistics study, we vary $n\in[1000, 10000]$ and set $k=10$, $\nicefrac{p}{q}=5$, and $\Bar{d}=100$.
We observe no significant performance fluctuations when varying other parameters.

\begin{figure}[!b]
\centering
\begin{tikzpicture}
\begin{groupplot}[group style={
                      group name=myplot,
                      group size= 3 by 2, horizontal sep=1.5cm,vertical sep=1cm},height=3cm,width=0.35\linewidth,title style={at={(0.5,0.9)},anchor=south},every axis x label/.style={at={(axis description cs:0.5,-0.15)},anchor=north},xmode=log,]
\nextgroupplot[
 	title =$\uparrow$ \textbf{Alg.\ Connectivity},
 	legend columns=4,
 	legend style={at={(0,1.4)},anchor=south west},
	legend cell align = left,
    legend entries={\textbf{\thiswork}, 1Tree, Edge Significance, Spectral Radius},
]
\addplot[very thick,color=cycle2] table[x=n,y=random_tree_algebraic_connectivity] {data/stats/sbm.tex};
\addplot[very thick,color=cycle4] table[x=n,y=random_plus_tree_algebraic_connectivity] {data/stats/sbm.tex};
\addplot[very thick,color=cycle5] table[x=n,y=mst_plus_best_filter_edge_significance_algebraic_connectivity] {data/stats/sbm.tex};
\addplot[very thick,color=cycle8] table[x=n,y=mst_plus_best_filter_spectral_radius_algebraic_connectivity] {data/stats/sbm.tex};

\nextgroupplot[
 	title =$\downarrow$ \textbf{Spectral radius},
]
\addplot[very thick,color=cycle2] table[x=n,y=random_tree_spectral_radius] {data/stats/sbm.tex};
\addplot[very thick,color=cycle4] table[x=n,y=random_plus_tree_spectral_radius] {data/stats/sbm.tex};
\addplot[very thick,color=cycle5] table[x=n,y=mst_plus_best_filter_edge_significance_spectral_radius] {data/stats/sbm.tex};
\addplot[very thick,color=cycle8] table[x=n,y=mst_plus_best_filter_spectral_radius_spectral_radius] {data/stats/sbm.tex};

\nextgroupplot[
 	title =$\downarrow$ \textbf{Eff.\ Resistance},
 	ymode=log
]
\addplot[very thick,color=cycle2] table[x=n,y=random_tree_effective_resistance] {data/stats/sbm.tex};
\addplot[very thick,color=cycle4] table[x=n,y=random_plus_tree_effective_resistance] {data/stats/sbm.tex};
\addplot[very thick,color=cycle5] table[x=n,y=mst_plus_best_filter_edge_significance_effective_resistance] {data/stats/sbm.tex};
\addplot[very thick,color=cycle8] table[x=n,y=mst_plus_best_filter_spectral_radius_effective_resistance] {data/stats/sbm.tex};

\nextgroupplot[
 	title =$\downarrow$ \textbf{Global CC},
 	ymode=log,
 	xlabel=$n$,
]
\addplot[very thick,color=cycle2] table[x=n,y=random_tree_global_clustering] {data/stats/sbm.tex};
\addplot[very thick,color=cycle4] table[x=n,y=random_plus_tree_global_clustering] {data/stats/sbm.tex};
\addplot[very thick,color=cycle5] table[x=n,y=mst_plus_best_filter_edge_significance_global_clustering] {data/stats/sbm.tex};
\addplot[very thick,color=cycle8] table[x=n,y=mst_plus_best_filter_spectral_radius_global_clustering] {data/stats/sbm.tex};

\nextgroupplot[
 	title =$\uparrow$ \textbf{log(\# trees)},
scaled y ticks = false,
    xlabel=$n$,
]
\addplot[very thick,color=cycle2] table[x=n,y=random_tree_number_of_trees] {data/stats/sbm.tex};
\addplot[very thick,color=cycle4] table[x=n,y=random_plus_tree_number_of_trees] {data/stats/sbm.tex};
\addplot[very thick,color=cycle5] table[x=n,y=mst_plus_best_filter_edge_significance_number_of_trees] {data/stats/sbm.tex};
\addplot[very thick,color=cycle8] table[x=n,y=mst_plus_best_filter_spectral_radius_number_of_trees] {data/stats/sbm.tex};

\nextgroupplot[
 	title =$\downarrow$ \textbf{log(\# triangles)},
 	xlabel=$n$,
]
\addplot[very thick,color=cycle2] table[x=n,y=random_tree_number_of_triangles] {data/stats/sbm.tex};
\addplot[very thick,color=cycle4] table[x=n,y=random_plus_tree_number_of_triangles] {data/stats/sbm.tex};
\addplot[very thick,color=cycle5] table[x=n,y=mst_plus_best_filter_edge_significance_number_of_triangles] {data/stats/sbm.tex};
\addplot[very thick,color=cycle8] table[x=n,y=mst_plus_best_filter_spectral_radius_number_of_triangles] {data/stats/sbm.tex};
\end{groupplot}
\end{tikzpicture}
\vspace*{-5mm}
\caption{Graph statistics measured on stochastic blockmodel graphs, averaged acros 1000 graphs with \nicefrac{p~}{q} ratio of 5, sparsified to average degree of 2. }\label{fig:stats-sbm}
\end{figure}
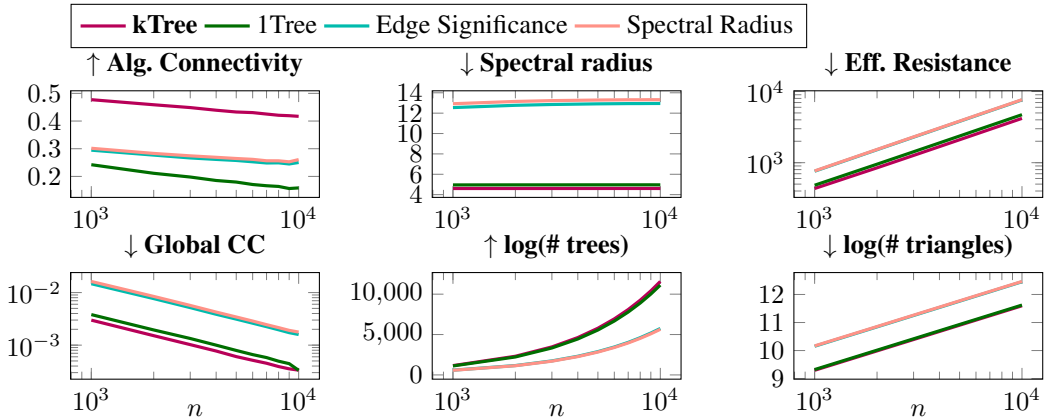 
\subsection{Graph Robustness Measures}\label{ssec:experiments-stats}

We evaluate five graph robustness measures from~\cite{chan2016optimizing} as well as two versions of the clustering coefficients of the graph.
For measures that require knowledge of all eigenvalues, we approximate the quantity via stochastic Lanczos quadrature method~\cite{ubaru2017fast} with 100 starting vectors and 10 iterations.
We provide a brief description of the measures, indicating whether a particular measure is ideally maximized ($\uparrow$) or minimized ($\downarrow$):
\begin{itemize}[noitemsep,topsep=2pt,parsep=0pt,partopsep=0pt]
    \item $\uparrow$ \textbf{Algebraic connectivity} is the smallest eigenvalue of the combinatorial graph Laplacian.
    \item $\downarrow$ \textbf{Spectral Radius} defined as the largest eigenvalue of the adjacency matrix of a graph.
    \item $\downarrow$ \textbf{Effective resistance} computed as $R=n\sum_i{\frac{1}{\lambda_i}}$.
    \item $\uparrow$ \textbf{Number of trees} computed\footnote{We omit the $log(n)$ normalization factor.} as $\log{S}=\sum_i{\lambda_i}$.
    \item $\downarrow$ \textbf{Number of triangles} computed as $\#\Delta=\frac{1}{6}\sum_i{\mu_i}$.
    \item $\downarrow$ \textbf{Global clustering coefficient}~\cite{newman2003structure} is defined as $\nicefrac{\tr{\mA^3}}{\sum_{i\neq{j}{\mA^2_{ij}}}}$.
    \item $\downarrow$ \textbf{Average local clustering coefficient}~\cite{watts1998collective} is defined as $c_i=\nicefrac{\sum_{j\in{}N_i}\sum_{k\in{}N(i)}\lvert{}e_{jk}\rvert}{d_i(d_i-1)}$. We average $c_i$ across all nodes in the graph.
\end{itemize}

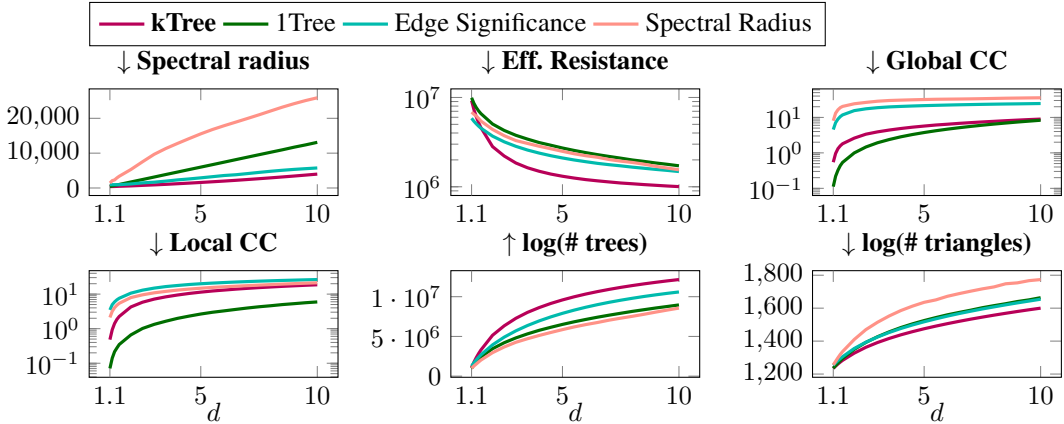
\begin{figure}[!t]
\centering
\begin{tikzpicture}
\begin{groupplot}[group style={
                      group name=myplot,
                      group size= 3 by 2, horizontal sep=1.5cm,vertical sep=1cm},height=3cm,width=0.35\linewidth,title style={at={(0.5,0.9)},anchor=south},every axis x label/.style={at={(axis description cs:0.5,-0.15)},anchor=north},extra x ticks={1.1},]
\nextgroupplot[
 	title =$\downarrow$ \textbf{Spectral radius},
    scaled y ticks = false,
 	legend columns=4,
 	legend style={at={(0,1.4)},anchor=south west},
	legend cell align = left,
    legend entries={\textbf{\thiswork}, 1Tree, Edge Significance, Spectral Radius},
]
\addplot[very thick,color=cycle2] table[x=degree,y=random-tree_spectral_radius] {data/stats/mnist_100.tex};
\addplot[very thick,color=cycle4] table[x=degree,y=random-plus-tree_spectral_radius] {data/stats/mnist_100.tex};
\addplot[very thick,color=cycle5] table[x=degree,y=mst-plus-best-filter-edge-significance_spectral_radius] {data/stats/mnist_100.tex};
\addplot[very thick,color=cycle8] table[x=degree,y=mst-plus-best-filter-spectral-radius_spectral_radius] {data/stats/mnist_100.tex};

\nextgroupplot[
 	title =$\downarrow$ \textbf{Eff.\ Resistance},
 	ymode=log,
]
\addplot[very thick,color=cycle2] table[x=degree,y=random-tree_effective_resistance] {data/stats/mnist_100.tex};
\addplot[very thick,color=cycle4] table[x=degree,y=random-plus-tree_effective_resistance] {data/stats/mnist_100.tex};
\addplot[very thick,color=cycle5] table[x=degree,y=mst-plus-best-filter-edge-significance_effective_resistance] {data/stats/mnist_100.tex};
\addplot[very thick,color=cycle8] table[x=degree,y=mst-plus-best-filter-spectral-radius_effective_resistance] {data/stats/mnist_100.tex};

\nextgroupplot[
 	title =$\downarrow$ \textbf{Global CC},
 	ymode=log,
]
\addplot[very thick,color=cycle2] table[x=degree,y=random-tree_global_clustering] {data/stats/mnist_100.tex};
\addplot[very thick,color=cycle4] table[x=degree,y=random-plus-tree_global_clustering] {data/stats/mnist_100.tex};
\addplot[very thick,color=cycle5] table[x=degree,y=mst-plus-best-filter-edge-significance_global_clustering] {data/stats/mnist_100.tex};
\addplot[very thick,color=cycle8] table[x=degree,y=mst-plus-best-filter-spectral-radius_global_clustering] {data/stats/mnist_100.tex};

\nextgroupplot[
 	title =$\downarrow$ \textbf{Local CC},
 	ymode=log,
 	xlabel={$d$},
]
\addplot[very thick,color=cycle2] table[x=degree,y=random-tree_local_clustering] {data/stats/mnist_100.tex};
\addplot[very thick,color=cycle4] table[x=degree,y=random-plus-tree_local_clustering] {data/stats/mnist_100.tex};
\addplot[very thick,color=cycle5] table[x=degree,y=mst-plus-best-filter-edge-significance_local_clustering] {data/stats/mnist_100.tex};
\addplot[very thick,color=cycle8] table[x=degree,y=mst-plus-best-filter-spectral-radius_local_clustering] {data/stats/mnist_100.tex};

\nextgroupplot[
 	title =$\uparrow$ \textbf{log(\# trees)},
scaled y ticks = false,
 	xlabel={$d$},
]
\addplot[very thick,color=cycle2] table[x=degree,y=random-tree_number_of_trees] {data/stats/mnist_100.tex};
\addplot[very thick,color=cycle4] table[x=degree,y=random-plus-tree_number_of_trees] {data/stats/mnist_100.tex};
\addplot[very thick,color=cycle5] table[x=degree,y=mst-plus-best-filter-edge-significance_number_of_trees] {data/stats/mnist_100.tex};
\addplot[very thick,color=cycle8] table[x=degree,y=mst-plus-best-filter-spectral-radius_number_of_trees] {data/stats/mnist_100.tex};

\nextgroupplot[
 	title =$\downarrow$ \textbf{log(\# triangles)},
 	xlabel={$d$},
]
\addplot[very thick,color=cycle2] table[x=degree,y=random-tree_number_of_triangles] {data/stats/mnist_100.tex};
\addplot[very thick,color=cycle4] table[x=degree,y=random-plus-tree_number_of_triangles] {data/stats/mnist_100.tex};
\addplot[very thick,color=cycle5] table[x=degree,y=mst-plus-best-filter-edge-significance_number_of_triangles] {data/stats/mnist_100.tex};
\addplot[very thick,color=cycle8] table[x=degree,y=mst-plus-best-filter-spectral-radius_number_of_triangles] {data/stats/mnist_100.tex};
\end{groupplot}
\end{tikzpicture}
\vspace*{-5mm}
\caption{Statistics measured on the $\varepsilon$-nearest-neighbor graph constructed from the {MNIST} dataset.}\label{fig:stats-mnist-maintext}
\end{figure}

We present results on the synthetic SBM graphs on Figure~\ref{fig:stats-sbm}.
Interestingly, the only metric with a critical difference between the \thiswork{} and 1Tree strategy is the algebraic connectivity of a graph.
Overall, we can observe a big difference between tree-based and greedy selection strategies, sometimes in the orders of magnitude better for random tree-based methods.

We present results on an exemplar MNIST graph on Figure~\ref{fig:stats-mnist-maintext}.
Figures for all other datasets can be found in Appendix.
There, we observe dramatic differences between approaches in terms of all of the metrics considered.
For real graphs, we do not report $\lambda_2$ because of numerical instabilities of finding it precisely in case when it is very close to 0.
Note how the differences in terms of the tree number are in logathmic terms, meaning \thiswork is better than the competitors by several orders of magnitude.
Compared to synthetic graphs, we observe stark contrast between different methods.

\subsection{Graph Clustering}\label{ssec:experiments-clustering}

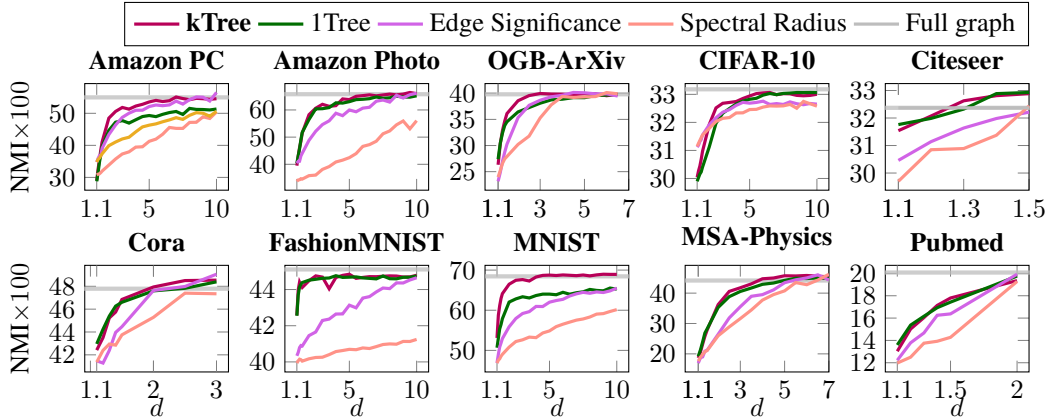
\begin{figure*}[t]
\centering
\begin{tikzpicture}
\begin{groupplot}[group style={
                      group name=myplot,
                      group size= 5 by 2, horizontal sep=0.75cm,vertical sep=1cm},height=3cm,width=0.25\linewidth,title style={at={(0.5,0.9)},anchor=south},every axis x label/.style={at={(axis description cs:0.5,-0.15)},anchor=north},extra x ticks={1.1},]
\nextgroupplot[
 	title = \textbf{Amazon PC},
	ylabel=NMI$\times100$,
]
\draw [color=cyclegray, ultra thick, draw opacity=0.75] (0,54.97) -- (12,54.97);
\addplot[very thick,color=cycle2] table[x=degree,y=random-tree_nmi] {data/clustering/amazon_electronics_computers.tex};
\addplot[very thick,color=cycle4] table[x=degree,y=random-plus-tree_nmi] {data/clustering/amazon_electronics_computers.tex};
\addplot[very thick,color=cycle6] table[x=degree,y=mst-plus-best-filter-edge-significance_nmi] {data/clustering/amazon_electronics_computers.tex};
\addplot[very thick,color=cycle8] table[x=degree,y=mst-plus-best-filter-spectral-radius_nmi] {data/clustering/amazon_electronics_computers.tex};
\addplot[very thick,color=cycle1] table[x=degree,y=random_nmi] {data/clustering/amazon_electronics_computers.tex};

\nextgroupplot[
 	title = \textbf{Amazon Photo},
]
\draw [color=cyclegray, ultra thick, draw opacity=0.75] (0,65.79) -- (12,65.79);
\addplot[very thick,color=cycle2] table[x=degree,y=random-tree_nmi] {data/clustering/amazon_electronics_photo.tex};
\addplot[very thick,color=cycle4] table[x=degree,y=random-plus-tree_nmi] {data/clustering/amazon_electronics_photo.tex};
\addplot[very thick,color=cycle6] table[x=degree,y=mst-plus-best-filter-edge-significance_nmi] {data/clustering/amazon_electronics_photo.tex};
\addplot[very thick,color=cycle8] table[x=degree,y=mst-plus-best-filter-spectral-radius_nmi] {data/clustering/amazon_electronics_photo.tex};

\nextgroupplot[
 	title = \textbf{OGB-ArXiv},
 	xmax=7,
 	xtick={1.1,3, 5, 7},
]
\draw [color=cyclegray, ultra thick, draw opacity=0.75] (0,39.80) -- (12,39.80);
\addplot[very thick,color=cycle2] table[x=degree,y=random-tree_nmi] {data/clustering/arxiv.tex};
\addplot[very thick,color=cycle4] table[x=degree,y=random-plus-tree_nmi] {data/clustering/arxiv.tex};
\addplot[very thick,color=cycle6] table[x=degree,y=mst-plus-best-filter-edge-significance_nmi] {data/clustering/arxiv.tex};
\addplot[very thick,color=cycle8] table[x=degree,y=mst-plus-best-filter-spectral-radius_nmi] {data/clustering/arxiv.tex};

\nextgroupplot[
 	title = \textbf{CIFAR-10},
]
\draw [color=cyclegray, ultra thick, draw opacity=0.75] (0,33.18) -- (12,33.18);
\addplot[very thick,color=cycle2] table[x=degree,y=random-tree_nmi] {data/clustering/cifar10_100.tex};
\addplot[very thick,color=cycle4] table[x=degree,y=random-plus-tree_nmi] {data/clustering/cifar10_100.tex};
\addplot[very thick,color=cycle6] table[x=degree,y=mst-plus-best-filter-edge-significance_nmi] {data/clustering/cifar10_100.tex};
\addplot[very thick,color=cycle8] table[x=degree,y=mst-plus-best-filter-spectral-radius_nmi] {data/clustering/cifar10_100.tex};

\nextgroupplot[
 	title = \textbf{Citeseer},
 	xmax=1.5,
 	xtick={1.1,1.3,1.5},
 	legend columns=6,
	legend style={at={(1,1.35)},anchor=south east},
    legend entries={\textbf{\thiswork}, 1Tree, Edge Significance, Spectral Radius, Full graph},
]
\draw [color=cyclegray, ultra thick, draw opacity=0.75] (0,32.37) -- (12,32.37);
\addplot[very thick,color=cycle2] table[x=degree,y=random-tree_nmi] {data/clustering/citeseer.tex};
\addplot[very thick,color=cycle4] table[x=degree,y=random-plus-tree_nmi] {data/clustering/citeseer.tex};
\addplot[very thick,color=cycle6] table[x=degree,y=mst-plus-best-filter-edge-significance_nmi] {data/clustering/citeseer.tex};
\addplot[very thick,color=cycle8] table[x=degree,y=mst-plus-best-filter-spectral-radius_nmi] {data/clustering/citeseer.tex};
\addplot[very thick,color=cyclegray] 
coordinates {
    (1.1,32.37)(1.5,32.37)
    };

\nextgroupplot[
 	title = \textbf{Cora},
 	xticklabels={,,},
 	extra x ticks={1.1, 2, 3},
	ylabel=NMI$\times100$,
	xlabel=$d$,
]
\draw [color=cyclegray, ultra thick, draw opacity=0.75] (0,47.79) -- (12,47.79);
\addplot[very thick,color=cycle2] table[x=degree,y=random-tree_nmi] {data/clustering/cora_full.tex};
\addplot[very thick,color=cycle4] table[x=degree,y=random-plus-tree_nmi] {data/clustering/cora_full.tex};
\addplot[very thick,color=cycle6] table[x=degree,y=mst-plus-best-filter-edge-significance_nmi] {data/clustering/cora_full.tex};
\addplot[very thick,color=cycle8] table[x=degree,y=mst-plus-best-filter-spectral-radius_nmi] {data/clustering/cora_full.tex};

\nextgroupplot[
 	title = \textbf{FashionMNIST},
	xlabel=$d$,
]
\draw [color=cyclegray, ultra thick, draw opacity=0.75] (0,45.11) -- (12,45.11);
\addplot[very thick,color=cycle2] table[x=degree,y=random-tree_nmi] {data/clustering/fashion_mnist_100.tex};
\addplot[very thick,color=cycle4] table[x=degree,y=random-plus-tree_nmi] {data/clustering/fashion_mnist_100.tex};
\addplot[very thick,color=cycle6] table[x=degree,y=mst-plus-best-filter-edge-significance_nmi] {data/clustering/fashion_mnist_100.tex};
\addplot[very thick,color=cycle8] table[x=degree,y=mst-plus-best-filter-spectral-radius_nmi] {data/clustering/fashion_mnist_100.tex};

\nextgroupplot[
 	title = \textbf{MNIST},
	xlabel=$d$,
]
\draw [color=cyclegray, ultra thick, draw opacity=0.75] (0,68.44) -- (12,68.44);
\addplot[very thick,color=cycle2] table[x=degree,y=random-tree_nmi] {data/clustering/mnist_100.tex};
\addplot[very thick,color=cycle4] table[x=degree,y=random-plus-tree_nmi] {data/clustering/mnist_100.tex};
\addplot[very thick,color=cycle6] table[x=degree,y=mst-plus-best-filter-edge-significance_nmi] {data/clustering/mnist_100.tex};
\addplot[very thick,color=cycle8] table[x=degree,y=mst-plus-best-filter-spectral-radius_nmi] {data/clustering/mnist_100.tex};

\nextgroupplot[
 	title = \textbf{MSA-Physics},
	xlabel=$d$,
 	xmax=7,
 	xtick={1.1,3, 5, 7},
]
\draw [color=cyclegray, ultra thick, draw opacity=0.75] (0,44.15) -- (12,44.15);
\addplot[very thick,color=cycle2] table[x=degree,y=random-tree_nmi] {data/clustering/ms_academic_phy.tex};
\addplot[very thick,color=cycle4] table[x=degree,y=random-plus-tree_nmi] {data/clustering/ms_academic_phy.tex};
\addplot[very thick,color=cycle6] table[x=degree,y=mst-plus-best-filter-edge-significance_nmi] {data/clustering/ms_academic_phy.tex};
\addplot[very thick,color=cycle8] table[x=degree,y=mst-plus-best-filter-spectral-radius_nmi] {data/clustering/ms_academic_phy.tex};

\nextgroupplot[
 	title = \textbf{Pubmed},
	xlabel=$d$,
]
\draw [color=cyclegray, ultra thick, draw opacity=0.75] (0,20.08) -- (12,20.07);
\addplot[very thick,color=cycle2] table[x=degree,y=random-tree_nmi] {data/clustering/pubmed.tex};
\addplot[very thick,color=cycle4] table[x=degree,y=random-plus-tree_nmi] {data/clustering/pubmed.tex};
\addplot[very thick,color=cycle6] table[x=degree,y=mst-plus-best-filter-edge-significance_nmi] {data/clustering/pubmed.tex};
\addplot[very thick,color=cycle8] table[x=degree,y=mst-plus-best-filter-spectral-radius_nmi] {data/clustering/pubmed.tex};
\end{groupplot}
\end{tikzpicture}
\vspace*{-6mm}
\caption{Clustering results on 10 real-world datasets. We vary the target average degree $d$ and report the normalized mutual information (multiplied by 100 for convenience) with respect to the ground-truth labels in each dataset. Random baseline is not present in this study due to the fact that disconnected components produce disconnected components that make NMI overly optimistic.}\label{fig:clustering}
\end{figure*}

We now discuss the performance of the graph clustering algorithms on sparsified graphs.
For each graph, we cluster it using the Louvain method~\cite{blondel2008fast} for community detection.
Figure~\ref{fig:clustering} reports the normalized mutual information between the clustering of the sparsified graph and ground-truth node labels on both natural and nearest neighbor graphs.

We observe that unweighted random tree-based methods produce significantly better results than their weighted counterparts regardless for both edge significance and spectral radius-based strategies\footnote{One might assume that there is an error in the weight calculation; however we have checked this thoroughly.}.
\thiswork{} is significantly better than 1Tree strategy on Amazon-PC, OGB-ArXiv, and MNIST datasets.
We can attribute that to the overall larger correlation of the label information to the ground-truth labels.
There is no case where it is losing to 1Tree.
In stark comparison, both weighting strategies of  \cite{chan2016optimizing,karhadkar2022fosr} and \cite{dianati2016unwinding} significantly underperform on all graphs we considered, with most degradation occurring in the very sparse regime.
This trend will continue in the other experiments, perhaps with a less severe trend: in general, we observe significant degradation of quality of all graph learning algorithms when using these sparsification techniques.
We do not report results of the completely random baseline, as it produces many disconnected components which get assigned a separate cluster, and NMI is ill-defined for these solutions.

Averaged across all datasets, the budget required for the best sparsification method to match the performance of graph clustering on the whole dataset is only 2--5 edges per node.
The only exception is Pubmed, where the graph structure seems to be very efficient, and all sparsification algorithms bring the performance down.

\subsection{Graph Embedding}\label{ssec:experiments-embedding}

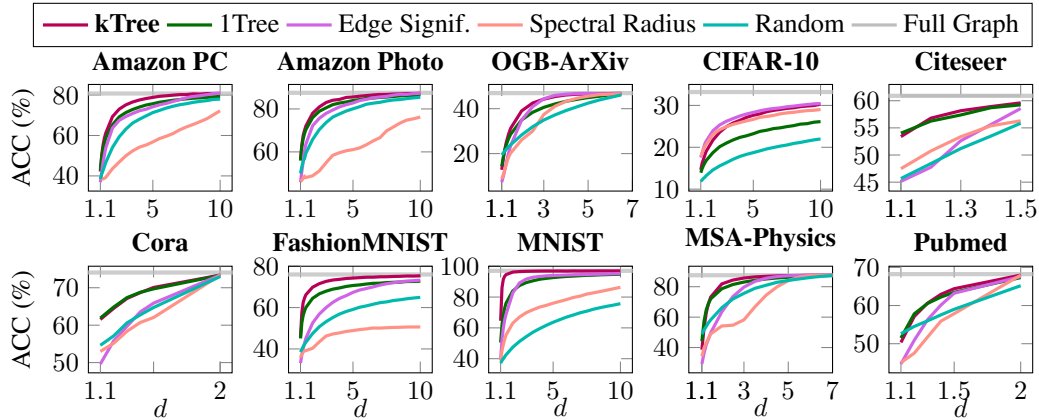
\begin{figure}[t]
\centering
\begin{tikzpicture}
\begin{groupplot}[group style={
                      group name=myplot,
                      group size= 5 by 2, horizontal sep=0.75cm,vertical sep=1cm},height=3cm,width=0.25\linewidth,title style={at={(0.5,0.9)},anchor=south},every axis x label/.style={at={(axis description cs:0.5,-0.15)},anchor=north},extra x ticks={1.1},]
\nextgroupplot[
 	title = \textbf{Amazon PC},
	ylabel=ACC (\%),
]
\draw [color=cyclegray, ultra thick, draw opacity=0.75] (0,80.75) -- (12,80.75);
\addplot[very thick,color=cycle2] table[x=degree,y=random-tree_ACC] {data/deepwalk/amazon_electronics_computers.tex};
\addplot[very thick,color=cycle4] table[x=degree,y=random-plus-tree_ACC] {data/deepwalk/amazon_electronics_computers.tex};
\addplot[very thick,color=cycle6] table[x=degree,y=mst-plus-best-filter-edge-significance_ACC] {data/deepwalk/amazon_electronics_computers.tex};
\addplot[very thick,color=cycle8] table[x=degree,y=mst-plus-best-filter-spectral-radius_ACC] {data/deepwalk/amazon_electronics_computers.tex};
\addplot[very thick,color=cycle5] table[x=degree,y=random_ACC] {data/deepwalk/amazon_electronics_computers.tex};

\nextgroupplot[
 	title = \textbf{Amazon Photo},
]
\draw [color=cyclegray, ultra thick, draw opacity=0.75] (0,87.52) -- (12,87.52);
\addplot[very thick,color=cycle2] table[x=degree,y=random-tree_ACC] {data/deepwalk/amazon_electronics_photo.tex};
\addplot[very thick,color=cycle4] table[x=degree,y=random-plus-tree_ACC] {data/deepwalk/amazon_electronics_photo.tex};
\addplot[very thick,color=cycle6] table[x=degree,y=mst-plus-best-filter-edge-significance_ACC] {data/deepwalk/amazon_electronics_photo.tex};
\addplot[very thick,color=cycle8] table[x=degree,y=mst-plus-best-filter-spectral-radius_ACC] {data/deepwalk/amazon_electronics_photo.tex};
\addplot[very thick,color=cycle5] table[x=degree,y=random_ACC] {data/deepwalk/amazon_electronics_photo.tex};

\nextgroupplot[
 	title = \textbf{OGB-ArXiv},
 	xmax=7,
 	xtick={1.1,3, 5, 7},
	]
\draw [color=cyclegray, ultra thick, draw opacity=0.75] (0,46.94) -- (12,46.94);
\addplot[very thick,color=cycle2] table[x=degree,y=random-tree_ACC] {data/deepwalk/arxiv.tex};
\addplot[very thick,color=cycle4] table[x=degree,y=random-plus-tree_ACC] {data/deepwalk/arxiv.tex};
\addplot[very thick,color=cycle6] table[x=degree,y=mst-plus-best-filter-edge-significance_ACC] {data/deepwalk/arxiv.tex};
\addplot[very thick,color=cycle8] table[x=degree,y=mst-plus-best-filter-spectral-radius_ACC] {data/deepwalk/arxiv.tex};
\addplot[very thick,color=cycle5] table[x=degree,y=random_ACC] {data/deepwalk/arxiv.tex};

\nextgroupplot[
 	title = \textbf{CIFAR-10},
 	ymax=35,
	]
\draw [color=cyclegray, ultra thick, draw opacity=0.75] (0,33.19) -- (12,33.19);
\addplot[very thick,color=cycle2] table[x=degree,y=random-tree_ACC] {data/deepwalk/cifar10_100.tex};
\addplot[very thick,color=cycle4] table[x=degree,y=random-plus-tree_ACC] {data/deepwalk/cifar10_100.tex};
\addplot[very thick,color=cycle6] table[x=degree,y=mst-plus-best-filter-edge-significance_ACC] {data/deepwalk/cifar10_100.tex};
\addplot[very thick,color=cycle8] table[x=degree,y=mst-plus-best-filter-spectral-radius_ACC] {data/deepwalk/cifar10_100.tex};
\addplot[very thick,color=cycle5] table[x=degree,y=random_ACC] {data/deepwalk/cifar10_100.tex};

\nextgroupplot[
 	title = \textbf{Citeseer},
 	xtick={1.1,1.3,1.5},
	ymax=63,
 	legend columns=6,
 	legend style={at={(1,1.35)},anchor=south east},
	legend cell align = left,
    legend entries={\textbf{\thiswork}, 1Tree, Edge Signif., Spectral Radius, Random, Full Graph},
]
\draw [color=cyclegray, ultra thick, draw opacity=0.75] (0,60.91) -- (12,60.91);
\addplot[very thick,color=cycle2] table[x=degree,y=random-tree_ACC] {data/deepwalk/citeseer.tex};
\addplot[very thick,color=cycle4] table[x=degree,y=random-plus-tree_ACC] {data/deepwalk/citeseer.tex};
\addplot[very thick,color=cycle6] table[x=degree,y=mst-plus-best-filter-edge-significance_ACC] {data/deepwalk/citeseer.tex};
\addplot[very thick,color=cycle8] table[x=degree,y=mst-plus-best-filter-spectral-radius_ACC] {data/deepwalk/citeseer.tex};
\addplot[very thick,color=cycle5] table[x=degree,y=random_ACC] {data/deepwalk/citeseer.tex};

\addplot[very thick,color=cyclegray] coordinates {
    (1.1,60.91)(1.5,60.91)
    };

\nextgroupplot[
 	title = \textbf{Cora},
 	xticklabels={,,},
 	extra x ticks={1.1, 2, 3},
	ylabel=ACC (\%),
	xlabel=$d$,
]
\draw [color=cyclegray, ultra thick, draw opacity=0.75] (0,74.06) -- (12,74.06);
\addplot[very thick,color=cycle2] table[x=degree,y=random-tree_ACC] {data/deepwalk/cora.tex};
\addplot[very thick,color=cycle4] table[x=degree,y=random-plus-tree_ACC] {data/deepwalk/cora.tex};
\addplot[very thick,color=cycle6] table[x=degree,y=mst-plus-best-filter-edge-significance_ACC] {data/deepwalk/cora.tex};
\addplot[very thick,color=cycle8] table[x=degree,y=mst-plus-best-filter-spectral-radius_ACC] {data/deepwalk/cora.tex};
\addplot[very thick,color=cycle5] table[x=degree,y=random_ACC] {data/deepwalk/cora.tex};

\nextgroupplot[
 	title = \textbf{FashionMNIST},
 	ymax=80,
	xlabel=$d$,
	]
\draw [color=cyclegray, ultra thick, draw opacity=0.75] (0,76.01) -- (12,76.01);
\addplot[very thick,color=cycle2] table[x=degree,y=random-tree_ACC] {data/deepwalk/fashion_mnist_100.tex};
\addplot[very thick,color=cycle4] table[x=degree,y=random-plus-tree_ACC] {data/deepwalk/fashion_mnist_100.tex};
\addplot[very thick,color=cycle6] table[x=degree,y=mst-plus-best-filter-edge-significance_ACC] {data/deepwalk/fashion_mnist_100.tex};
\addplot[very thick,color=cycle8] table[x=degree,y=mst-plus-best-filter-spectral-radius_ACC] {data/deepwalk/fashion_mnist_100.tex};
\addplot[very thick,color=cycle5] table[x=degree,y=random_ACC] {data/deepwalk/fashion_mnist_100.tex};

\nextgroupplot[
 	title = \textbf{MNIST},
	xlabel=$d$,
	ymax=100,
]
\draw [color=cyclegray, ultra thick, draw opacity=0.75] (0,97.02) -- (12,97.02);
\addplot[very thick,color=cycle2] table[x=degree,y=random-tree_ACC] {data/deepwalk/mnist_100.tex};
\addplot[very thick,color=cycle4] table[x=degree,y=random-plus-tree_ACC] {data/deepwalk/mnist_100.tex};
\addplot[very thick,color=cycle6] table[x=degree,y=mst-plus-best-filter-edge-significance_ACC] {data/deepwalk/mnist_100.tex};
\addplot[very thick,color=cycle8] table[x=degree,y=mst-plus-best-filter-spectral-radius_ACC] {data/deepwalk/mnist_100.tex};
\addplot[very thick,color=cycle5] table[x=degree,y=random_ACC] {data/deepwalk/mnist_100.tex};

\nextgroupplot[
 	title = \textbf{MSA-Physics},
 	xmax=7,
 	xtick={1.1,3, 5, 7},
	xlabel=$d$,
]
\draw [color=cyclegray, ultra thick, draw opacity=0.75] (0,87.55) -- (12,87.55);
\addplot[very thick,color=cycle2] table[x=degree,y=random-tree_ACC] {data/deepwalk/ms_academic_phy.tex};
\addplot[very thick,color=cycle4] table[x=degree,y=random-plus-tree_ACC] {data/deepwalk/ms_academic_phy.tex};
\addplot[very thick,color=cycle6] table[x=degree,y=mst-plus-best-filter-edge-significance_ACC] {data/deepwalk/ms_academic_phy.tex};
\addplot[very thick,color=cycle8] table[x=degree,y=mst-plus-best-filter-spectral-radius_ACC] {data/deepwalk/ms_academic_phy.tex};
\addplot[very thick,color=cycle5] table[x=degree,y=random_ACC] {data/deepwalk/ms_academic_phy.tex};

\nextgroupplot[
 	title = \textbf{Pubmed},
	xlabel=$d$,
]
\draw [color=cyclegray, ultra thick, draw opacity=0.75] (0,68.22) -- (12,68.22);
\addplot[very thick,color=cycle2] table[x=degree,y=random-tree_ACC] {data/deepwalk/pubmed.tex};
\addplot[very thick,color=cycle4] table[x=degree,y=random-plus-tree_ACC] {data/deepwalk/pubmed.tex};
\addplot[very thick,color=cycle6] table[x=degree,y=mst-plus-best-filter-edge-significance_ACC] {data/deepwalk/pubmed.tex};
\addplot[very thick,color=cycle8] table[x=degree,y=mst-plus-best-filter-spectral-radius_ACC] {data/deepwalk/pubmed.tex};
\addplot[very thick,color=cycle5] table[x=degree,y=random_ACC] {data/deepwalk/pubmed.tex};
\end{groupplot}
\end{tikzpicture}
\vspace*{-4mm}
\caption{Graph embedding performance on 10 real-world datasets. We vary the target average degree $d$ and report classification accuracy with respect to the ground-truth labels.}\label{fig:deepwalk}
\vspace*{-4mm}
\end{figure}

We now discuss the performance of graph embedding on sparsified graphs.
For each graph, we train a graph embedding~\cite{perozzi2014deepwalk} with parameters from the original paper (dimensionality 128, 80 walks per node of length 80, window size 10).
Then, we train a logistic regression model using scikit-learn~\cite{pedregosa2011scikit} with default parameters to predict the node labels.

Figure~\ref{fig:deepwalk} presents the results on 8 most informative datasets.
We observe that random tree-based methods are superior yet again, however, this time there is a noticeable difference in performance between \thiswork{} and 1Tree on almost all datasets.
We attribute that to the fact that DeepWalk algorithm performs aggressive smoothing of the input graph, so explicit decorrelation of the edges in the construction of \thiswork{} is more beneficial in this case.

Spectral radius-based weighting strategy is again performing the worst.
However, in the case of graph embedding, we can compare it to the random baseline: in 3 cases, it is significantly worse, in 2 it is better and in 3 more they are tied.
In this experiment, we can finally observe the extreme gains we can get by preserving the connectivity structure of graphs: the difference between the random baseline and \thiswork on MNIST dataset at its peak is more than 50\% in terms of accuracy!

\subsection{Graph Neural Networks}\label{ssec:experiments-gnns}

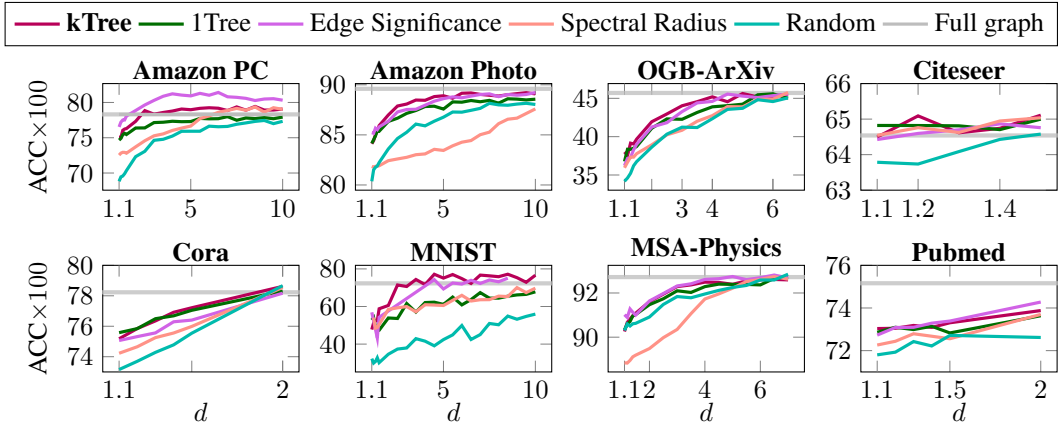
\begin{figure}[t]
\centering
\begin{tikzpicture}
\begin{groupplot}[group style={
                      group name=myplot,
                      group size= 4 by 2, horizontal sep=0.75cm,vertical sep=1cm},height=3cm,width=0.3\linewidth,title style={at={(0.5,0.8)},anchor=south},every axis x label/.style={at={(axis description cs:0.5,-0.2)},anchor=north},extra x ticks={1.1},]
\nextgroupplot[
 	title = \textbf{Amazon PC},
 	legend columns=6,
 	legend style={at={(-0.5,1.35)},anchor=south west},
	legend cell align = left,
    legend entries={\textbf{\thiswork}, 1Tree, Edge Significance, Spectral Radius, Random, Full graph},
	ylabel=ACC$\times100$,
]
\draw [color=cyclegray, ultra thick, draw opacity=0.75] (0,78.31) -- (12,78.31);
\addplot[very thick,color=cycle2] table[x=degree,y=random-tree] {data/gcn/amazon_electronics_computers.tex};
\addplot[very thick,color=cycle4] table[x=degree,y=random-plus-tree] {data/gcn/amazon_electronics_computers.tex};
\addplot[very thick,color=cycle6] table[x=degree,y=mst-plus-best-filter-edge-significance] {data/gcn/amazon_electronics_computers.tex};
\addplot[very thick,color=cycle8] table[x=degree,y=mst-plus-best-filter-spectral-radius] {data/gcn/amazon_electronics_computers.tex};
\addplot[very thick,color=cycle5] table[x=degree,y=random] {data/gcn/amazon_electronics_computers.tex};
\addplot[very thick,color=cyclegray] 
coordinates {
    (1.1,78.31)(10,78.31)
    };

\nextgroupplot[
 	title = \textbf{Amazon Photo},
]
\draw [color=cyclegray, ultra thick, draw opacity=0.75] (0,89.60) -- (12,89.60);
\addplot[very thick,color=cycle2] table[x=degree,y=random-tree] {data/gcn/amazon_electronics_photo.tex};
\addplot[very thick,color=cycle4] table[x=degree,y=random-plus-tree] {data/gcn/amazon_electronics_photo.tex};
\addplot[very thick,color=cycle6] table[x=degree,y=mst-plus-best-filter-edge-significance] {data/gcn/amazon_electronics_photo.tex};
\addplot[very thick,color=cycle8] table[x=degree,y=mst-plus-best-filter-spectral-radius] {data/gcn/amazon_electronics_photo.tex};
\addplot[very thick,color=cycle5] table[x=degree,y=random] {data/gcn/amazon_electronics_photo.tex};

\nextgroupplot[
 	title = \textbf{OGB-ArXiv},
 	xticklabels={,,},
 	extra x ticks={1.1, 3, 4, 6},
]
\draw [color=cyclegray, ultra thick, draw opacity=0.75] (0,45.70) -- (12,45.70);
\addplot[very thick,color=cycle2] table[x=degree,y=random-tree] {data/gcn/arxiv.tex};
\addplot[very thick,color=cycle4] table[x=degree,y=random-plus-tree] {data/gcn/arxiv.tex};
\addplot[very thick,color=cycle6] table[x=degree,y=mst-plus-best-filter-edge-significance] {data/gcn/arxiv.tex};
\addplot[very thick,color=cycle8] table[x=degree,y=mst-plus-best-filter-spectral-radius] {data/gcn/arxiv.tex};
\addplot[very thick,color=cycle5] table[x=degree,y=random] {data/gcn/arxiv.tex};

\nextgroupplot[
 	title = \textbf{Citeseer},
 	ymin=63,
 	ymax=66,
]
\draw [color=cyclegray, ultra thick, draw opacity=0.75] (0,64.54) -- (12,64.54);
\addplot[very thick,color=cycle2] table[x=degree,y=random-tree] {data/gcn/citeseer.tex};
\addplot[very thick,color=cycle4] table[x=degree,y=random-plus-tree] {data/gcn/citeseer.tex};
\addplot[very thick,color=cycle6] table[x=degree,y=mst-plus-best-filter-edge-significance] {data/gcn/citeseer.tex};
\addplot[very thick,color=cycle8] table[x=degree,y=mst-plus-best-filter-spectral-radius] {data/gcn/citeseer.tex};
\addplot[very thick,color=cycle5] table[x=degree,y=random] {data/gcn/citeseer.tex};

\nextgroupplot[
 	title = \textbf{Cora},
 	xticklabels={,,},
 	extra x ticks={1.1, 2, 3},
	ylabel=NMI$\times100$,
	ylabel=ACC$\times100$,
 	ymin=73,
 	ymax=80,
	xlabel=$d$,
]
\draw [color=cyclegray, ultra thick, draw opacity=0.75] (0,78.23) -- (12,78.23);
\addplot[very thick,color=cycle2] table[x=degree,y=random-tree] {data/gcn/cora.tex};
\addplot[very thick,color=cycle4] table[x=degree,y=random-plus-tree] {data/gcn/cora.tex};
\addplot[very thick,color=cycle6] table[x=degree,y=mst-plus-best-filter-edge-significance] {data/gcn/cora.tex};
\addplot[very thick,color=cycle8] table[x=degree,y=mst-plus-best-filter-spectral-radius] {data/gcn/cora.tex};
\addplot[very thick,color=cycle5] table[x=degree,y=random] {data/gcn/cora.tex};

\nextgroupplot[
 	title = \textbf{MNIST},
	xlabel=$d$,
]
\draw [color=cyclegray, ultra thick, draw opacity=0.75] (0,72.31) -- (12,72.31);
\addplot[very thick,color=cycle2] table[x=degree,y=random-tree] {data/gcn/mnist_100.tex};
\addplot[very thick,color=cycle4] table[x=degree,y=random-plus-tree] {data/gcn/mnist_100.tex};
\addplot[very thick,color=cycle6] table[x=degree,y=mst-plus-best-filter-edge-significance] {data/gcn/mnist_100.tex};
\addplot[very thick,color=cycle8] table[x=degree,y=mst-plus-best-filter-spectral-radius] {data/gcn/mnist_100.tex};
\addplot[very thick,color=cycle5] table[x=degree,y=random] {data/gcn/mnist_100.tex};

\nextgroupplot[
 	title = \textbf{MSA-Physics},
	xlabel=$d$,
	xlabel=$d$,
]
\draw [color=cyclegray, ultra thick, draw opacity=0.75] (0,92.72) -- (12,92.72);
\addplot[very thick,color=cycle2] table[x=degree,y=random-tree] {data/gcn/ms_academic_phy.tex};
\addplot[very thick,color=cycle4] table[x=degree,y=random-plus-tree] {data/gcn/ms_academic_phy.tex};
\addplot[very thick,color=cycle6] table[x=degree,y=mst-plus-best-filter-edge-significance] {data/gcn/ms_academic_phy.tex};
\addplot[very thick,color=cycle8] table[x=degree,y=mst-plus-best-filter-spectral-radius] {data/gcn/ms_academic_phy.tex};
\addplot[very thick,color=cycle5] table[x=degree,y=random] {data/gcn/ms_academic_phy.tex};

\nextgroupplot[
 	title = \textbf{Pubmed},
 	ymin=71 ,
 	ymax=76,
	xlabel=$d$,
]
\draw [color=cyclegray, ultra thick, draw opacity=0.75] (0,75.16) -- (12,75.16);
\addplot[very thick,color=cycle2] table[x=degree,y=random-tree] {data/gcn/pubmed.tex};
\addplot[very thick,color=cycle4] table[x=degree,y=random-plus-tree] {data/gcn/pubmed.tex};
\addplot[very thick,color=cycle6] table[x=degree,y=mst-plus-best-filter-edge-significance] {data/gcn/pubmed.tex};
\addplot[very thick,color=cycle8] table[x=degree,y=mst-plus-best-filter-spectral-radius] {data/gcn/pubmed.tex};
\addplot[very thick,color=cycle5] table[x=degree,y=random] {data/gcn/pubmed.tex};
\end{groupplot}
\end{tikzpicture}
\vspace*{-4mm}
\caption{GNN training results on 8 real-world datasets. We vary the target average degree $d$ and report the normalized mutual information (multiplied by 100 for convenience) with respect to the ground-truth labels in each dataset.}\label{fig:gcns}
\end{figure} 
We proceed with evaluating the performance of graph neural networks on sparsified graphs.
To unify the experimental setting across the 
For each graph, we train a basic Graph Convolutional Network (GCN) model~\cite{kipf2017semi} with 2 layers of 64 units each for 100 epochs.
We apply dropout to hidden units with a factor of 0.3 to stabilize the training process.

We present the results on Figure~\ref{fig:gcns}.
We can observe that on most datasets tree-based sparsification methods outperform other baselines.
Compared to graph clustering and embedding, graph neural networks are more robust to disconnected components---in fact, GNNs are less sensitive to structure of graphs overall, since these models have features to rely on.
Therefore, differences between methods are less pronounced for this graph learning approach.
However, we can still reap the benefits of tree-based sparsification: \thiswork{} is consistently a top performer.

We obtain sizeable benefits in sparsifying graphs for GNNs.
On all datasets, graph neural networks obtain performance comparable or better than the full graph at average degree equal to $\Bar{d}=5$, when this level of sparsification was available.
This point is obtained at slightly lower sparsity levels than for graph clustering and embedding, which can be explained by the fact that GNNs smooth the information via graph structure, and that process works best with more connections on average.

\subsection{General Observations and Trends}\label{ssec:general-trends}
Overall, our extensive experimental study suggests that finding very sparse \glottery winners is possible.
Our algorithms are able to offer significant improvements compared to baselines in terms of six graph structure quality metrics introduced in Section~\ref{sec:metrics}. 

On three distinct graph learning problems, we have showed that it is possible to obtain comparable \emph{or better} performance than the original graph structure with average node degree in the range 2--5.
Importantly, we show considerable performance improvements on graphs constructed from data.
 \section{Conclusion}\label{sec:conclusion}
This work postulates the \glottery hypothesis that states that extremely sparse backbones allow various graph learning algorithms to attain comparable performance as on the full graph.
We suggest two efficient algorithms to uncover such ``winning tickets''.
Our experimental results illustrate our methods' effectiveness, matching the performance of different graph learning algorithms in very sparse graphs ($\approx$~average degree of 5).
Extensions to bipartite graphs are of immediate interest since bipartite interaction graphs suffer from various problems with high-degree ``celebrity'' nodes. \fi

\iftrue
\cleardoublepage
\bibliographystyle{plain}
\bibliography{bibliography}
\fi

\iftrue
\newpage
\appendix
\onecolumn
\section{Appendix.}

\subsection{Dataset description}\label{sec:appendix-datasets}
Here we present a brief description of real-world datasets:
\begin{itemize}[noitemsep,topsep=0pt,parsep=0pt,partopsep=0pt]
    \item Cora, Citeseer, and Pubmed~\cite{sen2008} are citation networks; nodes represent papers connected by citation edges; features are bag-of-word abstracts, and labels represent paper topics. We use a re-processed version of Cora from~\cite{shchur2018pitfalls} due to errors in the processing of the original dataset.
    \item Amazon~\{PC, Photo\}~\cite{shchur2018pitfalls} are two subsets of the Amazon co-purchase graph for the computers and photo sections of the website, where nodes represent goods with edges between ones frequently purchased together; node features are bag-of-word reviews, and class labels are product category.
    \item OGB-ArXiv~\cite{hu2020open} is a paper co-citation dataset based on arXiv papers indexed by the Microsoft Academic graph. Nodes are papers; edges are citations, and class labels indicate the main category of the paper.
    \item CIFAR, MNIST, and FashionMNIST~\cite{krizhevsky2009learning,lecun1998mnist,xiao2017fashion} are $\varepsilon$-nearest neighbor graphs with $\varepsilon$ such that the average node degree is 100.
\end{itemize}

\begin{table}[!h]
\centering
\newcolumntype{C}{>{\raggedleft\arraybackslash}X}
\newcolumntype{S}{>{\centering\arraybackslash\hsize=.5\hsize}X}
\caption{\label{tbl:datasets}Dataset statistics. We report total number of nodes $|V|$, average node degree $\Bar{d}$, number of features $|X|$ and labels $|Y|$.}
\begin{tabularx}{\linewidth}{@{}p{1.85cm}CCCC@{}}
\toprule
\emph{dataset} & $|V|$ & $\Bar{d}$ & $|X|$ & $|Y|$ \\
\midrule
Cora & 19793 & 3.20 & 1433 & 7 \\
Citeseer & 3327 & 1.37 & 3703 & 6 \\
PubMed & 19717 & 2.25 & 500 & 3 \\
\mbox{Amazon PC} & 13752 & 17.88 & 767 & 10 \\
\mbox{Amazon Photo} & 7650 & 15.57 & 745 & 8 \\
\mbox{MSA-Physics} & 34493 & 7.19 & 8415 & 5 \\
\mbox{OGB-arXiv} & 169343 & 6.84 & 128 & 40 \\
CIFAR-10 & 50000 & 99 & 3072 & 10 \\
FashionMNIST & 60000 & 99 & 784 & 10 \\
MNIST & 60000 & 99 & 784 & 10 \\
\bottomrule
\end{tabularx}
\end{table}
 \cleardoublepage
\setcounter{topnumber}{10}
\setcounter{bottomnumber}{10}
\setcounter{totalnumber}{10}
\renewcommand{\topfraction}{1}
\renewcommand{\bottomfraction}{1}
\renewcommand{\textfraction}{0}
\renewcommand{\floatpagefraction}{1}
\subsection{Metrics on Real-World Datasets}
Here we present graph metrics computed on real-world graphs present in our experimental study.

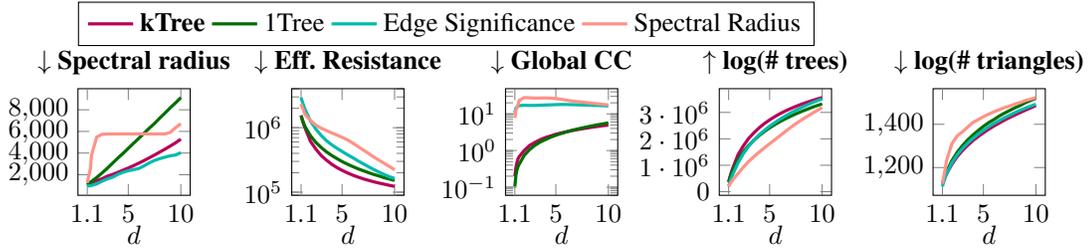
\begin{figure}[h]
\centering
\begin{tikzpicture}
\begin{groupplot}[group style={
                      group name=myplot,
                      group size= 5 by 1, horizontal sep=1.35cm,vertical sep=0.9cm},height=3cm,width=0.22\linewidth,title style={at={(0.5,0.9)},anchor=south},every axis x label/.style={at={(axis description cs:0.5,-0.2)},anchor=north},extra x ticks={1.1},]

\nextgroupplot[
 	title =$\downarrow$ \textbf{Spectral radius},
 	legend columns=4,
 	legend style={at={(0,1.4)},anchor=south west},
	legend cell align = left,
    legend entries={\textbf{\thiswork}, 1Tree, Edge Significance, Spectral Radius},
 	xlabel={$d$},
]
\addplot[very thick,color=cycle2] table[x=degree,y=random-tree_spectral_radius] {data/stats/amazon_electronics_computers.tex};
\addplot[very thick,color=cycle4] table[x=degree,y=random-plus-tree_spectral_radius] {data/stats/amazon_electronics_computers.tex};
\addplot[very thick,color=cycle5] table[x=degree,y=mst-plus-best-filter-edge-significance_spectral_radius] {data/stats/amazon_electronics_computers.tex};
\addplot[very thick,color=cycle8] table[x=degree,y=mst-plus-best-filter-spectral-radius_spectral_radius] {data/stats/amazon_electronics_computers.tex};

\nextgroupplot[
 	title =$\downarrow$ \textbf{Eff.\ Resistance},
 	ymode=log,
 	xlabel={$d$},
]
\addplot[very thick,color=cycle2] table[x=degree,y=random-tree_effective_resistance] {data/stats/amazon_electronics_computers.tex};
\addplot[very thick,color=cycle4] table[x=degree,y=random-plus-tree_effective_resistance] {data/stats/amazon_electronics_computers.tex};
\addplot[very thick,color=cycle5] table[x=degree,y=mst-plus-best-filter-edge-significance_effective_resistance] {data/stats/amazon_electronics_computers.tex};
\addplot[very thick,color=cycle8] table[x=degree,y=mst-plus-best-filter-spectral-radius_effective_resistance] {data/stats/amazon_electronics_computers.tex};

\nextgroupplot[
 	title =$\downarrow$ \textbf{Global CC},
 	ymode=log,
 	xlabel={$d$},
]
\addplot[very thick,color=cycle2] table[x=degree,y=random-tree_global_clustering] {data/stats/amazon_electronics_computers.tex};
\addplot[very thick,color=cycle4] table[x=degree,y=random-plus-tree_global_clustering] {data/stats/amazon_electronics_computers.tex};
\addplot[very thick,color=cycle5] table[x=degree,y=mst-plus-best-filter-edge-significance_global_clustering] {data/stats/amazon_electronics_computers.tex};
\addplot[very thick,color=cycle8] table[x=degree,y=mst-plus-best-filter-spectral-radius_global_clustering] {data/stats/amazon_electronics_computers.tex};

\nextgroupplot[
 	title =$\uparrow$ \textbf{log(\# trees)},
scaled y ticks = false,
 	xlabel={$d$},
]
\addplot[very thick,color=cycle2] table[x=degree,y=random-tree_number_of_trees] {data/stats/amazon_electronics_computers.tex};
\addplot[very thick,color=cycle4] table[x=degree,y=random-plus-tree_number_of_trees] {data/stats/amazon_electronics_computers.tex};
\addplot[very thick,color=cycle5] table[x=degree,y=mst-plus-best-filter-edge-significance_number_of_trees] {data/stats/amazon_electronics_computers.tex};
\addplot[very thick,color=cycle8] table[x=degree,y=mst-plus-best-filter-spectral-radius_number_of_trees] {data/stats/amazon_electronics_computers.tex};

\nextgroupplot[
 	title =$\downarrow$ \textbf{log(\# triangles)},
 	xlabel={$d$},
]
\addplot[very thick,color=cycle2] table[x=degree,y=random-tree_number_of_triangles] {data/stats/amazon_electronics_computers.tex};
\addplot[very thick,color=cycle4] table[x=degree,y=random-plus-tree_number_of_triangles] {data/stats/amazon_electronics_computers.tex};
\addplot[very thick,color=cycle5] table[x=degree,y=mst-plus-best-filter-edge-significance_number_of_triangles] {data/stats/amazon_electronics_computers.tex};
\addplot[very thick,color=cycle8] table[x=degree,y=mst-plus-best-filter-spectral-radius_number_of_triangles] {data/stats/amazon_electronics_computers.tex};
\end{groupplot}
\end{tikzpicture}
\vspace*{-4mm}
\caption{Graph statistics measured on the {AmazonPC} graph.}\label{fig:stats-amazon-pc}
\end{figure}
 \begin{figure}[h]
\centering
\begin{tikzpicture}
\begin{groupplot}[group style={
                      group name=myplot,
                      group size= 5 by 1, horizontal sep=1.35cm,vertical sep=0.9cm},height=3cm,width=0.22\linewidth,title style={at={(0.5,0.9)},anchor=south},every axis x label/.style={at={(axis description cs:0.5,-0.2)},anchor=north},extra x ticks={1.1},]

\nextgroupplot[
 	title =$\downarrow$ \textbf{Spectral radius},
 	legend columns=4,
 	legend style={at={(0,1.4)},anchor=south west},
	legend cell align = left,
    legend entries={\textbf{\thiswork}, 1Tree, Edge Significance, Spectral Radius},
 	xlabel={$d$},
]
\addplot[very thick,color=cycle2] table[x=degree,y=random-tree_spectral_radius] {data/stats/amazon_electronics_photo.tex};
\addplot[very thick,color=cycle4] table[x=degree,y=random-plus-tree_spectral_radius] {data/stats/amazon_electronics_photo.tex};
\addplot[very thick,color=cycle5] table[x=degree,y=mst-plus-best-filter-edge-significance_spectral_radius] {data/stats/amazon_electronics_photo.tex};
\addplot[very thick,color=cycle8] table[x=degree,y=mst-plus-best-filter-spectral-radius_spectral_radius] {data/stats/amazon_electronics_photo.tex};

\nextgroupplot[
 	title =$\downarrow$ \textbf{Eff.\ Resistance},
 	ymode=log,
 	xlabel={$d$},
]
\addplot[very thick,color=cycle2] table[x=degree,y=random-tree_effective_resistance] {data/stats/amazon_electronics_photo.tex};
\addplot[very thick,color=cycle4] table[x=degree,y=random-plus-tree_effective_resistance] {data/stats/amazon_electronics_photo.tex};
\addplot[very thick,color=cycle5] table[x=degree,y=mst-plus-best-filter-edge-significance_effective_resistance] {data/stats/amazon_electronics_photo.tex};
\addplot[very thick,color=cycle8] table[x=degree,y=mst-plus-best-filter-spectral-radius_effective_resistance] {data/stats/amazon_electronics_photo.tex};

\nextgroupplot[
 	title =$\downarrow$ \textbf{Global CC},
 	ymode=log,
 	xlabel={$d$},
]
\addplot[very thick,color=cycle2] table[x=degree,y=random-tree_global_clustering] {data/stats/amazon_electronics_photo.tex};
\addplot[very thick,color=cycle4] table[x=degree,y=random-plus-tree_global_clustering] {data/stats/amazon_electronics_photo.tex};
\addplot[very thick,color=cycle5] table[x=degree,y=mst-plus-best-filter-edge-significance_global_clustering] {data/stats/amazon_electronics_photo.tex};
\addplot[very thick,color=cycle8] table[x=degree,y=mst-plus-best-filter-spectral-radius_global_clustering] {data/stats/amazon_electronics_photo.tex};

\nextgroupplot[
 	title =$\uparrow$ \textbf{log(\# trees)},
scaled y ticks = false,
 	xlabel={$d$},
]
\addplot[very thick,color=cycle2] table[x=degree,y=random-tree_number_of_trees] {data/stats/amazon_electronics_photo.tex};
\addplot[very thick,color=cycle4] table[x=degree,y=random-plus-tree_number_of_trees] {data/stats/amazon_electronics_photo.tex};
\addplot[very thick,color=cycle5] table[x=degree,y=mst-plus-best-filter-edge-significance_number_of_trees] {data/stats/amazon_electronics_photo.tex};
\addplot[very thick,color=cycle8] table[x=degree,y=mst-plus-best-filter-spectral-radius_number_of_trees] {data/stats/amazon_electronics_photo.tex};

\nextgroupplot[
 	title =$\downarrow$ \textbf{log(\# triangles)},
 	xlabel={$d$},
]
\addplot[very thick,color=cycle2] table[x=degree,y=random-tree_number_of_triangles] {data/stats/amazon_electronics_photo.tex};
\addplot[very thick,color=cycle4] table[x=degree,y=random-plus-tree_number_of_triangles] {data/stats/amazon_electronics_photo.tex};
\addplot[very thick,color=cycle5] table[x=degree,y=mst-plus-best-filter-edge-significance_number_of_triangles] {data/stats/amazon_electronics_photo.tex};
\addplot[very thick,color=cycle8] table[x=degree,y=mst-plus-best-filter-spectral-radius_number_of_triangles] {data/stats/amazon_electronics_photo.tex};
\end{groupplot}
\end{tikzpicture}
\vspace*{-4mm}
\caption{Graph statistics measured on the {AmazonPhoto} graph.}\label{fig:stats-amazon-photo}
\end{figure}
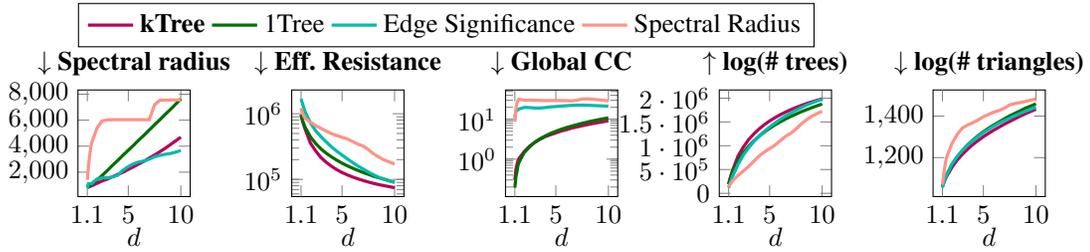
 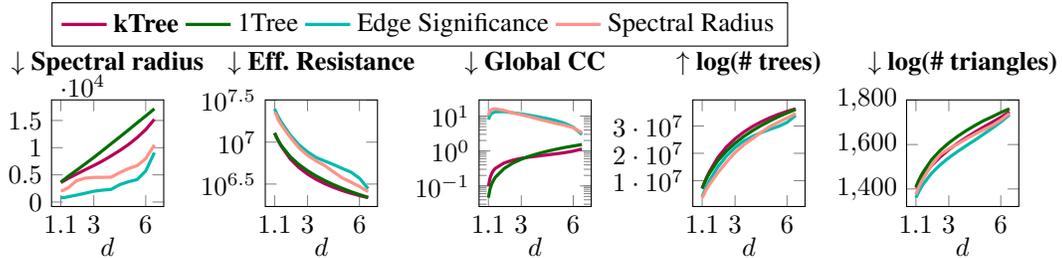
\begin{figure}[h]
\centering
\begin{tikzpicture}
\begin{groupplot}[group style={
                      group name=myplot,
                      group size= 5 by 1, horizontal sep=1.35cm,vertical sep=0.9cm},height=3cm,width=0.22\linewidth,title style={at={(0.5,1)},anchor=south},every axis x label/.style={at={(axis description cs:0.5,-0.2)},anchor=north},xtick={1.1, 3, 6}]

\nextgroupplot[
 	title =$\downarrow$ \textbf{Spectral radius},
 	legend columns=4,
 	legend style={at={(0,1.5)},anchor=south west},
	legend cell align = left,
    legend entries={\textbf{\thiswork}, 1Tree, Edge Significance, Spectral Radius},
 	xlabel={$d$},
]
\addplot[very thick,color=cycle2] table[x=degree,y=random-tree_spectral_radius] {data/stats/arxiv.tex};
\addplot[very thick,color=cycle4] table[x=degree,y=random-plus-tree_spectral_radius] {data/stats/arxiv.tex};
\addplot[very thick,color=cycle5] table[x=degree,y=mst-plus-best-filter-edge-significance_spectral_radius] {data/stats/arxiv.tex};
\addplot[very thick,color=cycle8] table[x=degree,y=mst-plus-best-filter-spectral-radius_spectral_radius] {data/stats/arxiv.tex};

\nextgroupplot[
 	title =$\downarrow$ \textbf{Eff.\ Resistance},
 	ymode=log,
 	xlabel={$d$},
]
\addplot[very thick,color=cycle2] table[x=degree,y=random-tree_effective_resistance] {data/stats/arxiv.tex};
\addplot[very thick,color=cycle4] table[x=degree,y=random-plus-tree_effective_resistance] {data/stats/arxiv.tex};
\addplot[very thick,color=cycle5] table[x=degree,y=mst-plus-best-filter-edge-significance_effective_resistance] {data/stats/arxiv.tex};
\addplot[very thick,color=cycle8] table[x=degree,y=mst-plus-best-filter-spectral-radius_effective_resistance] {data/stats/arxiv.tex};

\nextgroupplot[
 	title =$\downarrow$ \textbf{Global CC},
 	ymode=log,
 	xlabel={$d$},
]
\addplot[very thick,color=cycle2] table[x=degree,y=random-tree_global_clustering] {data/stats/arxiv.tex};
\addplot[very thick,color=cycle4] table[x=degree,y=random-plus-tree_global_clustering] {data/stats/arxiv.tex};
\addplot[very thick,color=cycle5] table[x=degree,y=mst-plus-best-filter-edge-significance_global_clustering] {data/stats/arxiv.tex};
\addplot[very thick,color=cycle8] table[x=degree,y=mst-plus-best-filter-spectral-radius_global_clustering] {data/stats/arxiv.tex};

\nextgroupplot[
 	title =$\uparrow$ \textbf{log(\# trees)},
scaled y ticks = false,
 	xlabel={$d$},
]
\addplot[very thick,color=cycle2] table[x=degree,y=random-tree_number_of_trees] {data/stats/arxiv.tex};
\addplot[very thick,color=cycle4] table[x=degree,y=random-plus-tree_number_of_trees] {data/stats/arxiv.tex};
\addplot[very thick,color=cycle5] table[x=degree,y=mst-plus-best-filter-edge-significance_number_of_trees] {data/stats/arxiv.tex};
\addplot[very thick,color=cycle8] table[x=degree,y=mst-plus-best-filter-spectral-radius_number_of_trees] {data/stats/arxiv.tex};

\nextgroupplot[
 	title =$\downarrow$ \textbf{log(\# triangles)},
 	xlabel={$d$},
]
\addplot[very thick,color=cycle2] table[x=degree,y=random-tree_number_of_triangles] {data/stats/arxiv.tex};
\addplot[very thick,color=cycle4] table[x=degree,y=random-plus-tree_number_of_triangles] {data/stats/arxiv.tex};
\addplot[very thick,color=cycle5] table[x=degree,y=mst-plus-best-filter-edge-significance_number_of_triangles] {data/stats/arxiv.tex};
\addplot[very thick,color=cycle8] table[x=degree,y=mst-plus-best-filter-spectral-radius_number_of_triangles] {data/stats/arxiv.tex};
\end{groupplot}
\end{tikzpicture}
\vspace*{-4mm}
\caption{Graph statistics measured on the {OGB-ArXiv} graph.}\label{fig:stats-arxiv}
\end{figure}
 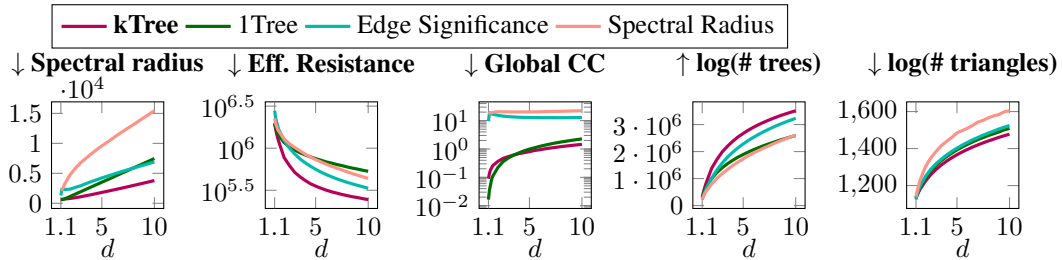
\begin{figure}[!h]
\centering
\begin{tikzpicture}
\begin{groupplot}[group style={
                      group name=myplot,
                      group size= 5 by 1, horizontal sep=1.35cm,vertical sep=0.9cm},height=3cm,width=0.22\linewidth,title style={at={(0.5,1)},anchor=south},every axis x label/.style={at={(axis description cs:0.5,-0.2)},anchor=north},extra x ticks={1.1},]

\nextgroupplot[
 	title =$\downarrow$ \textbf{Spectral radius},
 	legend columns=4,
 	legend style={at={(0,1.5)},anchor=south west},
	legend cell align = left,
    legend entries={\textbf{\thiswork}, 1Tree, Edge Significance, Spectral Radius},
 	xlabel={$d$},
]
\addplot[very thick,color=cycle2] table[x=degree,y=random-tree_spectral_radius] {data/stats/cifar10_100.tex};
\addplot[very thick,color=cycle4] table[x=degree,y=random-plus-tree_spectral_radius] {data/stats/cifar10_100.tex};
\addplot[very thick,color=cycle5] table[x=degree,y=mst-plus-best-filter-edge-significance_spectral_radius] {data/stats/cifar10_100.tex};
\addplot[very thick,color=cycle8] table[x=degree,y=mst-plus-best-filter-spectral-radius_spectral_radius] {data/stats/cifar10_100.tex};

\nextgroupplot[
 	title =$\downarrow$ \textbf{Eff.\ Resistance},
 	ymode=log,
 	xlabel={$d$},
]
\addplot[very thick,color=cycle2] table[x=degree,y=random-tree_effective_resistance] {data/stats/cifar10_100.tex};
\addplot[very thick,color=cycle4] table[x=degree,y=random-plus-tree_effective_resistance] {data/stats/cifar10_100.tex};
\addplot[very thick,color=cycle5] table[x=degree,y=mst-plus-best-filter-edge-significance_effective_resistance] {data/stats/cifar10_100.tex};
\addplot[very thick,color=cycle8] table[x=degree,y=mst-plus-best-filter-spectral-radius_effective_resistance] {data/stats/cifar10_100.tex};

\nextgroupplot[
 	title =$\downarrow$ \textbf{Global CC},
 	ymode=log,
 	xlabel={$d$},
]
\addplot[very thick,color=cycle2] table[x=degree,y=random-tree_global_clustering] {data/stats/cifar10_100.tex};
\addplot[very thick,color=cycle4] table[x=degree,y=random-plus-tree_global_clustering] {data/stats/cifar10_100.tex};
\addplot[very thick,color=cycle5] table[x=degree,y=mst-plus-best-filter-edge-significance_global_clustering] {data/stats/cifar10_100.tex};
\addplot[very thick,color=cycle8] table[x=degree,y=mst-plus-best-filter-spectral-radius_global_clustering] {data/stats/cifar10_100.tex};

\nextgroupplot[
 	title =$\uparrow$ \textbf{log(\# trees)},
scaled y ticks = false,
 	xlabel={$d$},
]
\addplot[very thick,color=cycle2] table[x=degree,y=random-tree_number_of_trees] {data/stats/cifar10_100.tex};
\addplot[very thick,color=cycle4] table[x=degree,y=random-plus-tree_number_of_trees] {data/stats/cifar10_100.tex};
\addplot[very thick,color=cycle5] table[x=degree,y=mst-plus-best-filter-edge-significance_number_of_trees] {data/stats/cifar10_100.tex};
\addplot[very thick,color=cycle8] table[x=degree,y=mst-plus-best-filter-spectral-radius_number_of_trees] {data/stats/cifar10_100.tex};

\nextgroupplot[
 	title =$\downarrow$ \textbf{log(\# triangles)},
 	xlabel={$d$},
]
\addplot[very thick,color=cycle2] table[x=degree,y=random-tree_number_of_triangles] {data/stats/cifar10_100.tex};
\addplot[very thick,color=cycle4] table[x=degree,y=random-plus-tree_number_of_triangles] {data/stats/cifar10_100.tex};
\addplot[very thick,color=cycle5] table[x=degree,y=mst-plus-best-filter-edge-significance_number_of_triangles] {data/stats/cifar10_100.tex};
\addplot[very thick,color=cycle8] table[x=degree,y=mst-plus-best-filter-spectral-radius_number_of_triangles] {data/stats/cifar10_100.tex};
\end{groupplot}
\end{tikzpicture}
\vspace*{-4mm}
\caption{Graph statistics measured on the {CIFAR10} graph.}\label{fig:stats-cifar10}
\end{figure}
 \begin{figure}[th]
\centering
\begin{tikzpicture}
\begin{groupplot}[group style={
                      group name=myplot,
                      group size= 5 by 1, horizontal sep=1.35cm,vertical sep=0.9cm},height=3cm,width=0.22\linewidth,title style={at={(0.5,0.9)},anchor=south},every axis x label/.style={at={(axis description cs:0.5,-0.2)},anchor=north},xtick={1.1, 2, 3},]

\nextgroupplot[
 	title =$\downarrow$ \textbf{Spectral radius},
 	legend columns=4,
 	legend style={at={(3.5,1.4)},anchor=south},
	legend cell align = left,
    legend entries={\textbf{\thiswork}, 1Tree, Edge Significance, Spectral Radius},
 	xlabel={$d$},
]
\addplot[very thick,color=cycle2] table[x=degree,y=random-tree_spectral_radius] {data/stats/cora_full.tex};
\addplot[very thick,color=cycle4] table[x=degree,y=random-plus-tree_spectral_radius] {data/stats/cora_full.tex};
\addplot[very thick,color=cycle5] table[x=degree,y=mst-plus-best-filter-edge-significance_spectral_radius] {data/stats/cora_full.tex};
\addplot[very thick,color=cycle8] table[x=degree,y=mst-plus-best-filter-spectral-radius_spectral_radius] {data/stats/cora_full.tex};

\nextgroupplot[
 	title =$\downarrow$ \textbf{Eff.\ Resistance},
 	ymode=log,
 	xlabel={$d$},
]
\addplot[very thick,color=cycle2] table[x=degree,y=random-tree_effective_resistance] {data/stats/cora_full.tex};
\addplot[very thick,color=cycle4] table[x=degree,y=random-plus-tree_effective_resistance] {data/stats/cora_full.tex};
\addplot[very thick,color=cycle5] table[x=degree,y=mst-plus-best-filter-edge-significance_effective_resistance] {data/stats/cora_full.tex};
\addplot[very thick,color=cycle8] table[x=degree,y=mst-plus-best-filter-spectral-radius_effective_resistance] {data/stats/cora_full.tex};

\nextgroupplot[
 	title =$\downarrow$ \textbf{Global CC},
 	ymode=log,
 	xlabel={$d$},
]
\addplot[very thick,color=cycle2] table[x=degree,y=random-tree_global_clustering] {data/stats/cora_full.tex};
\addplot[very thick,color=cycle4] table[x=degree,y=random-plus-tree_global_clustering] {data/stats/cora_full.tex};
\addplot[very thick,color=cycle5] table[x=degree,y=mst-plus-best-filter-edge-significance_global_clustering] {data/stats/cora_full.tex};
\addplot[very thick,color=cycle8] table[x=degree,y=mst-plus-best-filter-spectral-radius_global_clustering] {data/stats/cora_full.tex};

\nextgroupplot[
 	title =$\uparrow$ \textbf{log(\# trees)},
scaled y ticks = false,
 	xlabel={$d$},
]
\addplot[very thick,color=cycle2] table[x=degree,y=random-tree_number_of_trees] {data/stats/cora_full.tex};
\addplot[very thick,color=cycle4] table[x=degree,y=random-plus-tree_number_of_trees] {data/stats/cora_full.tex};
\addplot[very thick,color=cycle5] table[x=degree,y=mst-plus-best-filter-edge-significance_number_of_trees] {data/stats/cora_full.tex};
\addplot[very thick,color=cycle8] table[x=degree,y=mst-plus-best-filter-spectral-radius_number_of_trees] {data/stats/cora_full.tex};

\nextgroupplot[
 	title =$\downarrow$ \textbf{log(\# triangles)},
 	xlabel={$d$},
]
\addplot[very thick,color=cycle2] table[x=degree,y=random-tree_number_of_triangles] {data/stats/cora_full.tex};
\addplot[very thick,color=cycle4] table[x=degree,y=random-plus-tree_number_of_triangles] {data/stats/cora_full.tex};
\addplot[very thick,color=cycle5] table[x=degree,y=mst-plus-best-filter-edge-significance_number_of_triangles] {data/stats/cora_full.tex};
\addplot[very thick,color=cycle8] table[x=degree,y=mst-plus-best-filter-spectral-radius_number_of_triangles] {data/stats/cora_full.tex};
\end{groupplot}
\end{tikzpicture}
\caption{Graph statistics measured on the {Cora} graph.}\label{fig:stats-cora}
\end{figure}
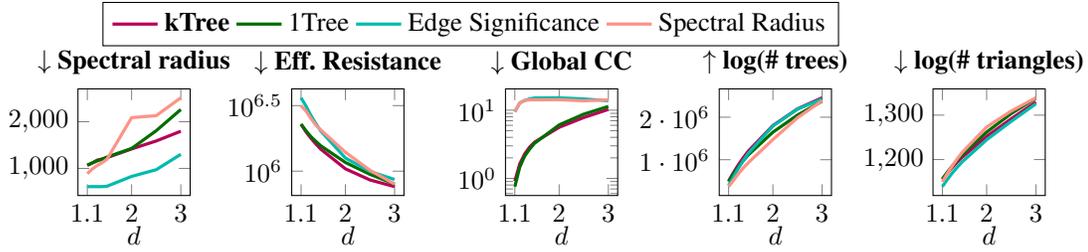
 \begin{figure}[h]
\centering
\begin{tikzpicture}
\begin{groupplot}[group style={
                      group name=myplot,
                      group size= 5 by 1, horizontal sep=1.35cm,vertical sep=0.9cm},height=3cm,width=0.22\linewidth,title style={at={(0.5,0.9)},anchor=south},every axis x label/.style={at={(axis description cs:0.5,-0.2)},anchor=north},extra x ticks={1.1},]

\nextgroupplot[
 	title =$\downarrow$ \textbf{Spectral radius},
 	legend columns=4,
 	legend style={at={(0,1.4)},anchor=south west},
	legend cell align = left,
    legend entries={\textbf{\thiswork}, 1Tree, Edge Significance, Spectral Radius},
 	xlabel={$d$},
    scaled y ticks = false,
]
\addplot[very thick,color=cycle2] table[x=degree,y=random-tree_spectral_radius] {data/stats/fashion_mnist_100.tex};
\addplot[very thick,color=cycle4] table[x=degree,y=random-plus-tree_spectral_radius] {data/stats/fashion_mnist_100.tex};
\addplot[very thick,color=cycle5] table[x=degree,y=mst-plus-best-filter-edge-significance_spectral_radius] {data/stats/fashion_mnist_100.tex};
\addplot[very thick,color=cycle8] table[x=degree,y=mst-plus-best-filter-spectral-radius_spectral_radius] {data/stats/fashion_mnist_100.tex};

\nextgroupplot[
 	title =$\downarrow$ \textbf{Eff.\ Resistance},
 	ymode=log,
 	xlabel={$d$},
]
\addplot[very thick,color=cycle2] table[x=degree,y=random-tree_effective_resistance] {data/stats/fashion_mnist_100.tex};
\addplot[very thick,color=cycle4] table[x=degree,y=random-plus-tree_effective_resistance] {data/stats/fashion_mnist_100.tex};
\addplot[very thick,color=cycle5] table[x=degree,y=mst-plus-best-filter-edge-significance_effective_resistance] {data/stats/fashion_mnist_100.tex};
\addplot[very thick,color=cycle8] table[x=degree,y=mst-plus-best-filter-spectral-radius_effective_resistance] {data/stats/fashion_mnist_100.tex};

\nextgroupplot[
 	title =$\downarrow$ \textbf{Global CC},
 	ymode=log,
 	xlabel={$d$},
]
\addplot[very thick,color=cycle2] table[x=degree,y=random-tree_global_clustering] {data/stats/fashion_mnist_100.tex};
\addplot[very thick,color=cycle4] table[x=degree,y=random-plus-tree_global_clustering] {data/stats/fashion_mnist_100.tex};
\addplot[very thick,color=cycle5] table[x=degree,y=mst-plus-best-filter-edge-significance_global_clustering] {data/stats/fashion_mnist_100.tex};
\addplot[very thick,color=cycle8] table[x=degree,y=mst-plus-best-filter-spectral-radius_global_clustering] {data/stats/fashion_mnist_100.tex};

\nextgroupplot[
 	title =$\uparrow$ \textbf{log(\# trees)},
scaled y ticks = false,
 	xlabel={$d$},
]
\addplot[very thick,color=cycle2] table[x=degree,y=random-tree_number_of_trees] {data/stats/fashion_mnist_100.tex};
\addplot[very thick,color=cycle4] table[x=degree,y=random-plus-tree_number_of_trees] {data/stats/fashion_mnist_100.tex};
\addplot[very thick,color=cycle5] table[x=degree,y=mst-plus-best-filter-edge-significance_number_of_trees] {data/stats/fashion_mnist_100.tex};
\addplot[very thick,color=cycle8] table[x=degree,y=mst-plus-best-filter-spectral-radius_number_of_trees] {data/stats/fashion_mnist_100.tex};

\nextgroupplot[
 	title =$\downarrow$ \textbf{log(\# triangles)},
 	xlabel={$d$},
]
\addplot[very thick,color=cycle2] table[x=degree,y=random-tree_number_of_triangles] {data/stats/fashion_mnist_100.tex};
\addplot[very thick,color=cycle4] table[x=degree,y=random-plus-tree_number_of_triangles] {data/stats/fashion_mnist_100.tex};
\addplot[very thick,color=cycle5] table[x=degree,y=mst-plus-best-filter-edge-significance_number_of_triangles] {data/stats/fashion_mnist_100.tex};
\addplot[very thick,color=cycle8] table[x=degree,y=mst-plus-best-filter-spectral-radius_number_of_triangles] {data/stats/fashion_mnist_100.tex};
\end{groupplot}
\end{tikzpicture}
\vspace*{-4mm}
\caption{Graph statistics measured on the {FashionMNIST} graph.}\label{fig:stats-fashion-mnist}
\end{figure}
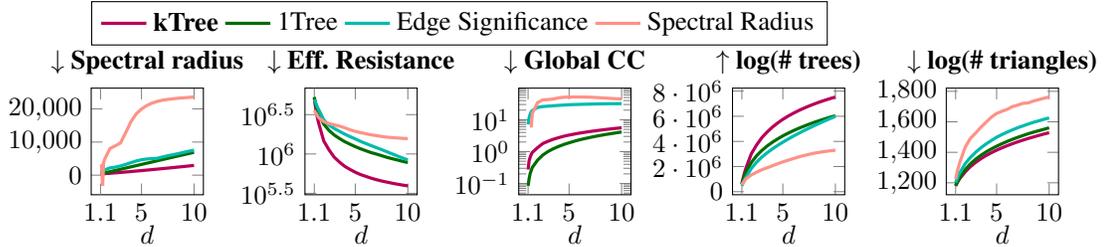
 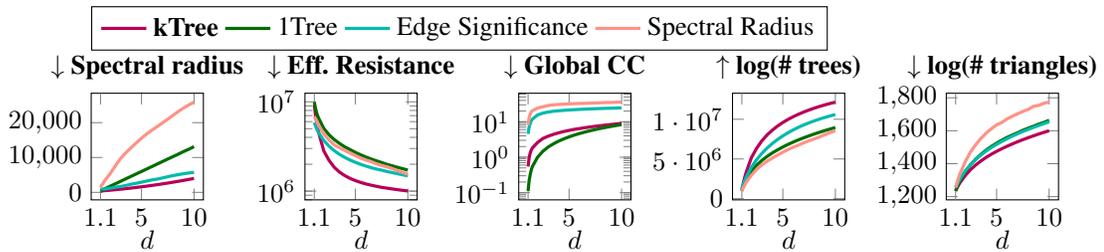
\begin{figure}[h]
\centering
\begin{tikzpicture}
\begin{groupplot}[group style={
                      group name=myplot,
                      group size= 5 by 1, horizontal sep=1.35cm,vertical sep=0.9cm},height=3cm,width=0.22\linewidth,title style={at={(0.5,0.9)},anchor=south},every axis x label/.style={at={(axis description cs:0.5,-0.2)},anchor=north},extra x ticks={1.1},]

\nextgroupplot[
 	title =$\downarrow$ \textbf{Spectral radius},
 	legend columns=4,
 	legend style={at={(0,1.4)},anchor=south west},
	legend cell align = left,
    legend entries={\textbf{\thiswork}, 1Tree, Edge Significance, Spectral Radius},
 	xlabel={$d$},
    scaled y ticks = false,
]
\addplot[very thick,color=cycle2] table[x=degree,y=random-tree_spectral_radius] {data/stats/mnist_100.tex};
\addplot[very thick,color=cycle4] table[x=degree,y=random-plus-tree_spectral_radius] {data/stats/mnist_100.tex};
\addplot[very thick,color=cycle5] table[x=degree,y=mst-plus-best-filter-edge-significance_spectral_radius] {data/stats/mnist_100.tex};
\addplot[very thick,color=cycle8] table[x=degree,y=mst-plus-best-filter-spectral-radius_spectral_radius] {data/stats/mnist_100.tex};

\nextgroupplot[
 	title =$\downarrow$ \textbf{Eff.\ Resistance},
 	ymode=log,
 	xlabel={$d$},
]
\addplot[very thick,color=cycle2] table[x=degree,y=random-tree_effective_resistance] {data/stats/mnist_100.tex};
\addplot[very thick,color=cycle4] table[x=degree,y=random-plus-tree_effective_resistance] {data/stats/mnist_100.tex};
\addplot[very thick,color=cycle5] table[x=degree,y=mst-plus-best-filter-edge-significance_effective_resistance] {data/stats/mnist_100.tex};
\addplot[very thick,color=cycle8] table[x=degree,y=mst-plus-best-filter-spectral-radius_effective_resistance] {data/stats/mnist_100.tex};

\nextgroupplot[
 	title =$\downarrow$ \textbf{Global CC},
 	ymode=log,
 	xlabel={$d$},
]
\addplot[very thick,color=cycle2] table[x=degree,y=random-tree_global_clustering] {data/stats/mnist_100.tex};
\addplot[very thick,color=cycle4] table[x=degree,y=random-plus-tree_global_clustering] {data/stats/mnist_100.tex};
\addplot[very thick,color=cycle5] table[x=degree,y=mst-plus-best-filter-edge-significance_global_clustering] {data/stats/mnist_100.tex};
\addplot[very thick,color=cycle8] table[x=degree,y=mst-plus-best-filter-spectral-radius_global_clustering] {data/stats/mnist_100.tex};

\nextgroupplot[
 	title =$\uparrow$ \textbf{log(\# trees)},
scaled y ticks = false,
 	xlabel={$d$},
]
\addplot[very thick,color=cycle2] table[x=degree,y=random-tree_number_of_trees] {data/stats/mnist_100.tex};
\addplot[very thick,color=cycle4] table[x=degree,y=random-plus-tree_number_of_trees] {data/stats/mnist_100.tex};
\addplot[very thick,color=cycle5] table[x=degree,y=mst-plus-best-filter-edge-significance_number_of_trees] {data/stats/mnist_100.tex};
\addplot[very thick,color=cycle8] table[x=degree,y=mst-plus-best-filter-spectral-radius_number_of_trees] {data/stats/mnist_100.tex};

\nextgroupplot[
 	title =$\downarrow$ \textbf{log(\# triangles)},
 	xlabel={$d$},
]
\addplot[very thick,color=cycle2] table[x=degree,y=random-tree_number_of_triangles] {data/stats/mnist_100.tex};
\addplot[very thick,color=cycle4] table[x=degree,y=random-plus-tree_number_of_triangles] {data/stats/mnist_100.tex};
\addplot[very thick,color=cycle5] table[x=degree,y=mst-plus-best-filter-edge-significance_number_of_triangles] {data/stats/mnist_100.tex};
\addplot[very thick,color=cycle8] table[x=degree,y=mst-plus-best-filter-spectral-radius_number_of_triangles] {data/stats/mnist_100.tex};
\end{groupplot}
\end{tikzpicture}
\vspace*{-4mm}
\caption{Graph statistics measured on the {MNIST} graph.}\label{fig:stats-mnist}
\end{figure}
 \begin{figure}[h]
\centering
\begin{tikzpicture}
\begin{groupplot}[group style={
                      group name=myplot,
                      group size= 5 by 1, horizontal sep=1.35cm,vertical sep=0.9cm},height=3cm,width=0.22\linewidth,title style={at={(0.5,0.9)},anchor=south},every axis x label/.style={at={(axis description cs:0.5,-0.2)},anchor=north},xtick={1.1, 3, 6}]

\nextgroupplot[
 	title =$\downarrow$ \textbf{Spectral radius},
 	legend columns=4,
 	legend style={at={(0,1.4)},anchor=south west},
	legend cell align = left,
    legend entries={\textbf{\thiswork}, 1Tree, Edge Significance, Spectral Radius},
 	xlabel={$d$},
]
\addplot[very thick,color=cycle2] table[x=degree,y=random-tree_spectral_radius] {data/stats/ms_academic_phy.tex};
\addplot[very thick,color=cycle4] table[x=degree,y=random-plus-tree_spectral_radius] {data/stats/ms_academic_phy.tex};
\addplot[very thick,color=cycle5] table[x=degree,y=mst-plus-best-filter-edge-significance_spectral_radius] {data/stats/ms_academic_phy.tex};
\addplot[very thick,color=cycle8] table[x=degree,y=mst-plus-best-filter-spectral-radius_spectral_radius] {data/stats/ms_academic_phy.tex};

\nextgroupplot[
 	title =$\downarrow$ \textbf{Eff.\ Resistance},
 	ymode=log,
 	xlabel={$d$},
]
\addplot[very thick,color=cycle2] table[x=degree,y=random-tree_effective_resistance] {data/stats/ms_academic_phy.tex};
\addplot[very thick,color=cycle4] table[x=degree,y=random-plus-tree_effective_resistance] {data/stats/ms_academic_phy.tex};
\addplot[very thick,color=cycle5] table[x=degree,y=mst-plus-best-filter-edge-significance_effective_resistance] {data/stats/ms_academic_phy.tex};
\addplot[very thick,color=cycle8] table[x=degree,y=mst-plus-best-filter-spectral-radius_effective_resistance] {data/stats/ms_academic_phy.tex};

\nextgroupplot[
 	title =$\downarrow$ \textbf{Global CC},
 	ymode=log,
 	xlabel={$d$},
]
\addplot[very thick,color=cycle2] table[x=degree,y=random-tree_global_clustering] {data/stats/ms_academic_phy.tex};
\addplot[very thick,color=cycle4] table[x=degree,y=random-plus-tree_global_clustering] {data/stats/ms_academic_phy.tex};
\addplot[very thick,color=cycle5] table[x=degree,y=mst-plus-best-filter-edge-significance_global_clustering] {data/stats/ms_academic_phy.tex};
\addplot[very thick,color=cycle8] table[x=degree,y=mst-plus-best-filter-spectral-radius_global_clustering] {data/stats/ms_academic_phy.tex};

\nextgroupplot[
 	title =$\uparrow$ \textbf{log(\# trees)},
scaled y ticks = false,
 	xlabel={$d$},
]
\addplot[very thick,color=cycle2] table[x=degree,y=random-tree_number_of_trees] {data/stats/ms_academic_phy.tex};
\addplot[very thick,color=cycle4] table[x=degree,y=random-plus-tree_number_of_trees] {data/stats/ms_academic_phy.tex};
\addplot[very thick,color=cycle5] table[x=degree,y=mst-plus-best-filter-edge-significance_number_of_trees] {data/stats/ms_academic_phy.tex};
\addplot[very thick,color=cycle8] table[x=degree,y=mst-plus-best-filter-spectral-radius_number_of_trees] {data/stats/ms_academic_phy.tex};

\nextgroupplot[
 	title =$\downarrow$ \textbf{log(\# triangles)},
 	xlabel={$d$},
]
\addplot[very thick,color=cycle2] table[x=degree,y=random-tree_number_of_triangles] {data/stats/ms_academic_phy.tex};
\addplot[very thick,color=cycle4] table[x=degree,y=random-plus-tree_number_of_triangles] {data/stats/ms_academic_phy.tex};
\addplot[very thick,color=cycle5] table[x=degree,y=mst-plus-best-filter-edge-significance_number_of_triangles] {data/stats/ms_academic_phy.tex};
\addplot[very thick,color=cycle8] table[x=degree,y=mst-plus-best-filter-spectral-radius_number_of_triangles] {data/stats/ms_academic_phy.tex};
\end{groupplot}
\end{tikzpicture}
\vspace*{-4mm}
\caption{Graph statistics measured on the {MSA-Physics} graph.}\label{fig:stats-ms-phy}
\end{figure}
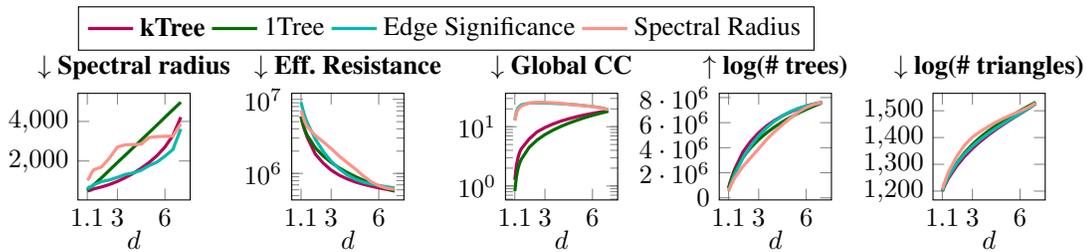
 \begin{figure}[h]
\centering
\begin{tikzpicture}
\begin{groupplot}[group style={
                      group name=myplot,
                      group size= 5 by 1, horizontal sep=1.35cm,vertical sep=0.9cm},height=3cm,width=0.22\linewidth,title style={at={(0.5,0.9)},anchor=south},every axis x label/.style={at={(axis description cs:0.5,-0.2)},anchor=north},extra x ticks={1.1},]

\nextgroupplot[
 	title =$\downarrow$ \textbf{Spectral radius},
 	legend columns=4,
 	legend style={at={(0,1.4)},anchor=south west},
	legend cell align = left,
    legend entries={\textbf{\thiswork}, 1Tree, Edge Significance, Spectral Radius},
 	xlabel={$d$},
]
\addplot[very thick,color=cycle2] table[x=degree,y=random-tree_spectral_radius] {data/stats/pubmed.tex};
\addplot[very thick,color=cycle4] table[x=degree,y=random-plus-tree_spectral_radius] {data/stats/pubmed.tex};
\addplot[very thick,color=cycle5] table[x=degree,y=mst-plus-best-filter-edge-significance_spectral_radius] {data/stats/pubmed.tex};
\addplot[very thick,color=cycle8] table[x=degree,y=mst-plus-best-filter-spectral-radius_spectral_radius] {data/stats/pubmed.tex};

\nextgroupplot[
 	title =$\downarrow$ \textbf{Eff.\ Resistance},
 	ymode=log,
 	xlabel={$d$},
]
\addplot[very thick,color=cycle2] table[x=degree,y=random-tree_effective_resistance] {data/stats/pubmed.tex};
\addplot[very thick,color=cycle4] table[x=degree,y=random-plus-tree_effective_resistance] {data/stats/pubmed.tex};
\addplot[very thick,color=cycle5] table[x=degree,y=mst-plus-best-filter-edge-significance_effective_resistance] {data/stats/pubmed.tex};
\addplot[very thick,color=cycle8] table[x=degree,y=mst-plus-best-filter-spectral-radius_effective_resistance] {data/stats/pubmed.tex};

\nextgroupplot[
 	title =$\downarrow$ \textbf{Global CC},
 	ymode=log,
 	xlabel={$d$},
]
\addplot[very thick,color=cycle2] table[x=degree,y=random-tree_global_clustering] {data/stats/pubmed.tex};
\addplot[very thick,color=cycle4] table[x=degree,y=random-plus-tree_global_clustering] {data/stats/pubmed.tex};
\addplot[very thick,color=cycle5] table[x=degree,y=mst-plus-best-filter-edge-significance_global_clustering] {data/stats/pubmed.tex};
\addplot[very thick,color=cycle8] table[x=degree,y=mst-plus-best-filter-spectral-radius_global_clustering] {data/stats/pubmed.tex};

\nextgroupplot[
 	title =$\uparrow$ \textbf{log(\# trees)},
scaled y ticks = false,
 	xlabel={$d$},
]
\addplot[very thick,color=cycle2] table[x=degree,y=random-tree_number_of_trees] {data/stats/pubmed.tex};
\addplot[very thick,color=cycle4] table[x=degree,y=random-plus-tree_number_of_trees] {data/stats/pubmed.tex};
\addplot[very thick,color=cycle5] table[x=degree,y=mst-plus-best-filter-edge-significance_number_of_trees] {data/stats/pubmed.tex};
\addplot[very thick,color=cycle8] table[x=degree,y=mst-plus-best-filter-spectral-radius_number_of_trees] {data/stats/pubmed.tex};

\nextgroupplot[
 	title =$\downarrow$ \textbf{log(\# triangles)},
 	xlabel={$d$},
]
\addplot[very thick,color=cycle2] table[x=degree,y=random-tree_number_of_triangles] {data/stats/pubmed.tex};
\addplot[very thick,color=cycle4] table[x=degree,y=random-plus-tree_number_of_triangles] {data/stats/pubmed.tex};
\addplot[very thick,color=cycle5] table[x=degree,y=mst-plus-best-filter-edge-significance_number_of_triangles] {data/stats/pubmed.tex};
\addplot[very thick,color=cycle8] table[x=degree,y=mst-plus-best-filter-spectral-radius_number_of_triangles] {data/stats/pubmed.tex};
\end{groupplot}
\end{tikzpicture}
\vspace*{-4mm}
\caption{Graph statistics measured on the {Pubmed} graph.}\label{fig:stats-pubmed}
\end{figure}
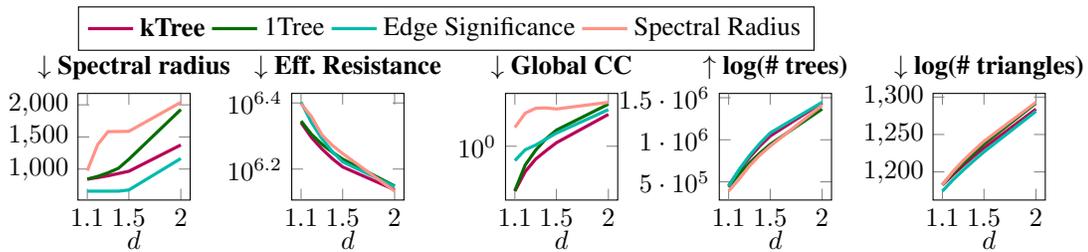
 \fi \end{document}